\documentclass[preprint, 5p,times]{elsarticle} 

\usepackage{graphics} 
\usepackage{epsfig} 
\usepackage{float}
\usepackage{subcaption}
\usepackage{hyperref}
\usepackage{siunitx}
\usepackage{amsmath} 
\usepackage{amssymb}  
\usepackage{graphicx} 
\usepackage{cite}
\usepackage{algorithm,algorithmic}
\usepackage{booktabs}
\usepackage{multirow}
\usepackage{amsmath}
\usepackage{xcolor}
\usepackage{caption}
\usepackage{siunitx}
\usepackage{url}
\usepackage{nidanfloat}
\usepackage{lipsum}
\usepackage{adjustbox}

\newcommand*{\tran}{^{\mkern-1.5mu\mathsf{T}}}

\begin{document}
\begin{frontmatter}

\title{Motion-based Camera Localization System in Colonoscopy Videos}

\author[add1]{Heming Yao\corref{cor1}}
\ead{hemingy@umich.edu}
\author[add2]{Ryan W. Stidham}
\ead{ryanstid@med.umich.edu}
\author[add1]{Zijun Gao}
\ead{zijung@umich.edu}
\author[add1,add3]{Jonathan Gryak}
\ead{gryakj@med.umich.edu}
\author[add1,add3,add4,add5]{Kayvan Najarian}
\ead{kayvan@med.umich.edu}

\cortext[cor1]{Corresponding author}
\address[add1]{Department of Computational Medicine and Bioinformatics, University of Michigan, Ann Arbor, MI, USA}
\address[add2]{Department of Gastroenterology, University of Michigan, Ann Arbor, MI, USA}
\address[add3]{Michigan Institute for Data Science, University of Michigan, Ann Arbor, MI, USA}
\address[add4]{Michigan Center for Integrative Research in Critical Care, University of Michigan, Ann Arbor, MI, USA}
\address[add5]{Department of Emergency Medicine, University of Michigan, Ann Arbor, MI, USA}

\begin{abstract}
Optical colonoscopy is an essential diagnostic and prognostic tool for many gastrointestinal diseases, including cancer screening and staging, intestinal bleeding, diarrhea, abdominal symptom evaluation, and inflammatory bowel disease assessment. However, the evaluation, classification, and quantification of findings from colonoscopy are subject to inter-observer variation. Automated assessment of colonoscopy is of interest considering the subjectivity present in qualitative human interpretations of colonoscopy findings. Localization of the camera is essential to interpreting the meaning and context of findings for diseases evaluated by colonoscopy. In this study, we propose a camera localization system to estimate the relative location of the camera and classify the colon into anatomical segments. The camera localization system begins with non-informative frame detection and removal. Then a self-training end-to-end convolutional neural network is built to estimate the camera motion, where several strategies are proposed to improve its robustness and generalization on endoscopic videos. Using the estimated camera motion a camera trajectory can be derived and a relative location index calculated. Based on the estimated location index, anatomical colon segment classification is performed by constructing a colon template. The proposed motion estimation algorithm was evaluated on an external dataset containing the ground truth for camera pose. The experimental results show that the performance of the proposed method is superior to other published methods. The relative location index estimation and anatomical region classification were further validated using colonoscopy videos collected from routine clinical practice. This validation yielded an average accuracy in classification of 0.754, which is substantially higher than the performances obtained using location indices built from other methods.

\end{abstract}

\begin{keyword}
Colonoscopy Video Analysis, Camera Motion Estimation, Endoscopy, Localization
\end{keyword}
\end{frontmatter}

\section{INTRODUCTION}
Optical colonoscopy is a medical procedure to inspect the mucosal surface to detect abnormalities in the colon. It is an indispensable tool for evaluating many gastrointestinal diseases.  During the procedure, a flexible probe with a charge-coupled device (CCD) camera and a fiber optic light source at the tip are inserted into the rectum and advanced through the colon.  After reaching the proximal portion of the intestine (typically the cecum or ileum), the physician withdraws the colonoscope and visually inspects the tissue surface for abnormalities.  Colono-scopy is the primary tool used to screen for colorectal cancer and precancerous lesions and is recommended for all adults over age 50, with repeated exams every 3-10 years based on risk stratification \citep{rex2017colorectal, us2008screening}.  Additionally, colonoscopy is commonly used to locate and treat sources of lower gastrointestinal bleeding.
\citep{farrell2005management, chaudhry1998colonoscopy}.  Colonoscopy is also used to investigate causes of diarrhea, in particular inflammatory bowel diseases (IBD) \citep{shah2001usefulness}.  Inspection of the mucosal surface by colonoscopy is particularly important in IBD, where disease severity assessment and monitoring of treatment effectiveness are heavily dependent on the findings of repeated colonoscopies \citep{rubin2019acg, peyrin2015selecting, food19ulcerative}. However, the interpretation of medical images and video are subject to inter-observer variation, limiting the reliability, quantification, and tracking of the evolution of findings from colonoscopy over time.  Thus, there is a need to automate the analysis of colonoscopies in order to standardize reporting so that disease assessments are uniform regardless of colonoscope operator experience.

Though image classification has been the focus of efforts in automated colonoscopy video analysis, camera localization is an important component for interpreting findings. Localizing lesions can help generate contextual information by providing anatomical awareness.  The localization of the camera can help improve the accuracy in diagnosis and prognosis of multiple diseases identified by colonoscopy, including cancer position and burden, bleeding source, and IBD distribution \citep{cotton2008practical,derwinger2011variations,steinberg1986prognostic,aziz2017paediatric}. Existing methods for endoscopic object localization fall into two categories: sensor-based localization methods \citep{than2012review} and computer vision-based camera motion estimation methods\citep{armin2016automated, liu2013optical}. While sensor-based methods can provide the absolute position, a relative position of the camera with respect to the end and start of the colon may be sufficient for contextual understanding of colon features. Sensor-free methods of camera location estimation using computer-vision methods are likely to sacrifice marginally relevant accuracy in exchange for improved feasibility.  Sensor-free camera localization methods are attractive as there is no additional equipment needed, allowing for rapid integration into clinical workflows, wide availability, and low financial cost.

In this work, we propose a vision-based sensor-free localization system using deep learning techniques.  Deep learning architectures have been successfully applied to estimating camera pose in driving videos \citep{zhou2017unsupervised, liu2015learning, wang2018recurrent}. In \citep{szeliski1999prediction}, a new view can be synthesized given the current view coupled with predicted depth and camera motion under the assumption of a rigid scene. In \citep{zhou2017unsupervised}, a self-supervised convolutional nerual network (CNN) architecture was proposed for camera motion estimation with the model trained using the photometric error between the synthesized view and the real frame. From their results, the model can achieve performance comparable with state-of-the-art methods without knowing the ground truth for camera motion. In this study, we adapted the self-supervised framework and proposed several modifications to improve the performance of pose estimation in colonoscopy videos. Based on the camera motion estimation, a localization system was developed that can identify the location index of each frame as well as which anatomical colon segment it belongs to. An overview of the system is shown in Figure \ref{fig:overview}. The proposed localization system starts with the detection and removal of non-informative frames and frames with biopsy forceps. The non-informative frames are removed as they contain no motion information and may hamper optical flow calculation and pose estimation. The calculation of photometric error for self-supervision relies on the assumption of a rigid scene, which is not true during a biopsy. As a result, frames with biopsy forceps are removed. After that, a self-training end-to-end CNN framework estimates the camera motion within colonoscopy videos. After camera motion estimation, the camera trajectory is derived, from which the location index estimation and anatomical colon segment classification are estimated. The output from the proposed motion-based localization system can further facilitate contextual understanding, such as determining the distribution of biopsy sites and disease severity. 

The main contributions of our work are as follows:

\begin{figure}[] 
\includegraphics[width=3.6in]{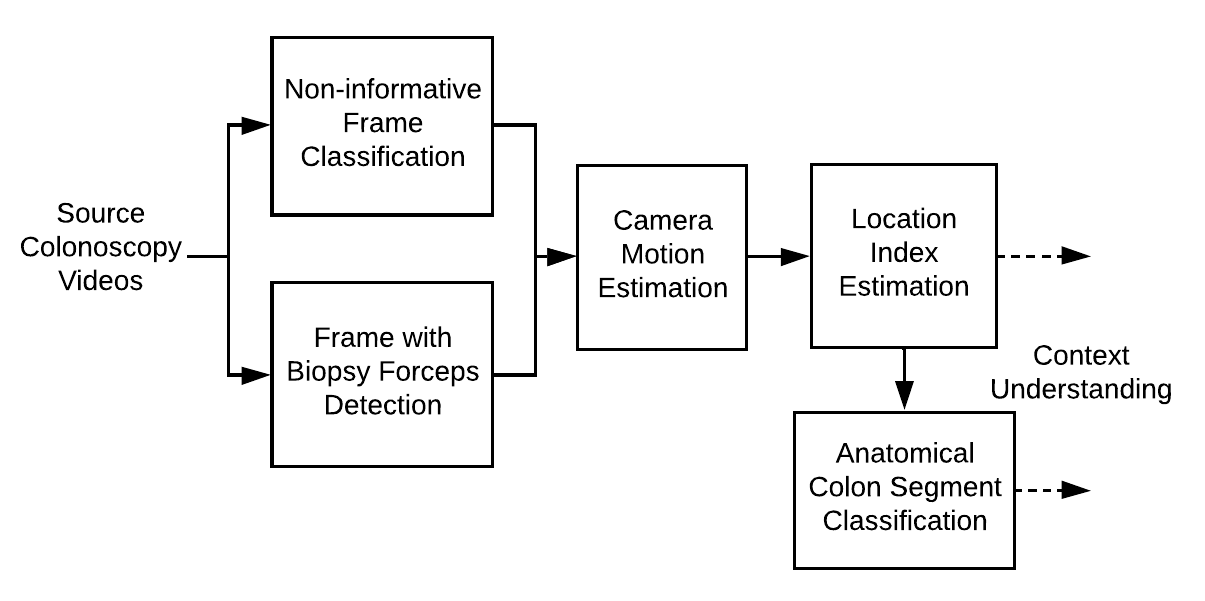}
\caption{An overview of the proposed localization system for colonoscopy videos.}
\label{fig:overview}
\end{figure}

\begin{enumerate}
  \item A novel camera localization system was designed that overcomes challenges of motion tracking in colonoscopy videos. Three major modifications were made over previous methods to improve the performance of pose estimation in endoscopic videos. First, optical flow between two consecutive frames was added as input, which provides explicit information regarding the camera's motion and reduces the impact of video noise. Secondly, a mask for specular regions was estimated to filter out regions not compatible with the Lambertian surface assumption. In colonoscopy videos, moisture on the colon wall will lead to a number of specular regions, which may disrupt photometric error calculation. Filtering out those regions focuses the network on the photometric error calculated from the Lambertian surface. Thirdly, a movement consistency term was added to improve the robustness of the network. The EndoSLAM dataset \citep{ozyoruk2020quantitative}, an external dataset with ground truth for camera pose, was used to validate the proposed modifications. The proposed camera pose estimation network achieved the best results on endoscopic videos compared with other published methods.
 
  \item A novel anatomical colon segment classification algorithm was devised. Utilizing the estimated location index, a colon segment template that represents the relative length of each anatomical colon segment over the entire colon is constructed. With a new colonoscopy video, anatomical colon segment classification can be performed by estimating the location index and referencing the constructed colon template. The colon segment can not only be used to provide anatomical contextual information but also to evaluate the accuracy of the localization system.
  
  \item The proposed localization system was applied to colon-oscopy videos collected from routine practice and validated using colon segments manually annotated by experienced gastroenterologists. The performance of the colon segment classification was compared with other methods. An additional metric was calculated using data from ScopeGuide$^{\textrm{\textregistered}}$ (Olympus Corporation, \url{https://www.olympus-global.com/}), which can provide approximate length information of the inserted scope into the colon. The location index derived from ScopeGuide length was compared with the location index estimated from the proposed localization system. To the best of the authors' knowledge, it is the first time that a localization system has been proposed and evaluated on colonoscopy videos from routine practice. 
\end{enumerate}

\section{Related Work}
\label{sec:related_work}
\subsection{Self-supervised learning for camera pose estimation on monocular videos}
Deep learning architectures have achieved success in relative camera pose estimation and single view depth estimation \citep{liu2015learning,ummenhofer2017demon, wang2020flow} on monocular videos. While traditional camera pose estimation algorithms, such as visual odometry, are effective in certain settings, their reliance on accurate image correspondence matching causes problems when the images are of low texture and from a complex environment. Deep learning methods may overcome these challenges by additional supervision. 

Based on the view synthesis technique \citep{szeliski1999prediction}, the photometric error between the synthesized new view and real new view can be used as supervision to train the network, which eliminates the requirement for ground truth data. In \citep{zhou2017unsupervised}, an end-to-end framework named SfMLearner was proposed to jointly train the single view depth and camera pose using projection error. This framework utilized unlabeled data but had a performance comparable with approaches that require ground truth. Authors in \citep{wang2018learning} proposed a simple normalization of the estimated depth map, which can effectively avoid depth prediction saturating to zero. In addition, a Direct Visual Odometry (DVO) \citep{steinbrucker2011real} pose predictor was incorporated into the end-to-end training to establish a direct relationship between the depth map and the camera pose prediction. Joint pose and depth estimations prevent the determination of absolute scale. To recover the absolute scale, UnDeepVO, proposed in \citep{li2018undeepvo}, was trained using stereo image pairs. After training, the model was tested on consecutive monocular images, yielding good performance on pose estimation.

In the self-supervised framework using photometric loss, several assumptions are implicitly made, including (1) a static scene; (2) Lambertian reflectance, i.e., the brightness is constant regardless of the observer's viewing angle; (3) no change in lighting between two consecutive frames; and (4) no occlusion between two consecutive frames. These assumptions may fail in real-world applications. A number of studies proposed additional loss terms to improve the robustness of the network in these circumstances. In \citep{zhan2018unsupervised}, in addition to photometric loss, deep feature-based warping loss was proposed to take contextual information into consideration rather than per-pixel color matching alone. Salient feature correspondences were extracted in \citep{shen2019beyond}, and a matching loss constrained by epipolar geometry was proposed to improve network optimization. GeoNet was proposed in \citep{yin2018geonet} to jointly estimate monocular depth, camera pose, and optical flow. A geometric consistency measurement was proposed as an additional loss term to improve the network's resilience to outliers. The GeoNet model achieved state-of-art performance on the KITTI dataset for all three tasks. In \citep{babu2018undemon}, the self-supervised framework was extended by applying the Charbonnier penalty to combine spatial and temporal reconstruction losses. In \citep{bian2019unsupervised}, a geometry consistency constraint was proposed to enforce the scale-consistency of depth and pose networks. Their results showed that the proposed pose network achieves performance commensurate with methods using stereo videos. In \citep{hosseinzadeh2020unsupervised}, a re-estimation approach was proposed where the camera pose estimation was decomposed into a sequence of smaller pose estimation problems. For smaller pose estimation problems, the assumptions made in camera pose estimation algorithms are more likely to hold.

\subsection{Vision-based camera pose estimation in monocular endoscopic videos}
Camera motion tracking in endoscopic videos has been investigated in a few studies previously. An optical flow approach to tracking colonoscopy video was proposed in \citep{liu2013optical}. The focus of expansion (FOE) was calculated using a combination of sparse and dense optical flow calculations. Based on calculated FOE, the computation of the camera's rotation and translation parameters from the optical flow field were separated. The depth of frame was estimated from a tube-like model. This approach is sensitive to optical flow and FOE calculations. In \citep{armin2016automated}, Kanade-Lucas-Tomasi features were extracted and tracked through consecutive frames. After that, a visual odometry algorithm was used to calculate camera motion by assuming each pair of consecutive frames to be a stereo pair. In this approach, the translation parameters estimated from visual odometry are subject to arbitrary scaling. The camera's speed has to be used to facilitate camera motion estimation, which makes it somewhat impractical. 
Deep learning has also been applied to camera pose estimation for endoscopic videos. A convolutional neural network was proposed in \citep{armin2017learning} to estimate the pose of the colonoscope. The network was trained and evaluated on simulated videos. Their results showed that the pose estimation from CNN was more accurate and faster than feature-based computer vision methods. In \citep{turan2018deep}, the depth map was first estimated from images using the Tsai–Shah Shape from Shading method \citep{ping1994shape}. After that, a recurrent convolutional neural network was applied to model dynamics across the frames and estimate the camera pose. While the evaluation on a real pig stomach dataset showed that the method achieved high translational and rotational accuracy, ground truth for camera pose was also required to train the network. A self-supervised framework has also been applied to endoscopic videos. In \citep{turan2018unsupervised}, SfMLeaner was applied to estimate the motion of the endoscopic capsule robot. Evaluations on videos collected from ex-vivo porcine stomach were used to demonstrate the effectiveness of the method. In \citep{freedman2020detecting}, a calibration-free framework was proposed by estimating the camera intrinsic parameters. In \citep{li2020unsupervised}, temporal information among consecutive frames was explored by a long-short-term-memory layer to improve the accuracy of pose estimation. 

\subsection{Summary and discussion}
Recent progress in self-supervised camera pose estimation has focused on extending the framework and loss function to improve the network's robustness to violations of the assumptions made by photometric loss. The majority of the methods were trained and validated on the KITTI dataset \citep{geiger2013vision}, which has fewer rotational variations. The self-supervised framework proposed for driving videos has already been successfully applied to endoscopic videos. However, one important limitation of existing work is that they were only validated on either simulated endoscopic videos or a sequence of selected frames from real videos.

The aim of this study is to track the camera within colono-scopy videos. Camera motion tracking in the colon environment can be very challenging because of its complex geometry and a low and similar textural pattern across the colon. From a previous study, a deep learning-based motion estimation method is more suitable than a feature matching based system \citep{turan2018deep}. In this study, the self-learning framework of \citep{turan2018deep} was extended to estimate the dense depth map and camera pose jointly. For colonoscopy videos, the motion between two consecutive frames is usually quite small. As such, one can assume the scene is static and lighting changes minimal in the absence of biopsies. However, in the colon environment, the assumption of a Lambertian surface may not hold due to the presence of surface moisture. Specular regions may exist where the brightness can change significantly under different viewing angles. The existence of specular regions impairs the calculation of photometric loss and that of other previously proposed loss terms such as image similarity loss \citep{yin2018geonet} and feature matching-based loss\citep{zhan2018unsupervised}.

To overcome this challenge, a specular mask is estimated by extending the network to correct the photometric loss. Additionally, optical flow was added as input and a calculated motion consistency term was used to to improve the robustness and generalization of the network. As the intrinsic complexity of the colon environment poses several challenges to camera motion estimation, it is necessary to validate the proposed method using real colonoscopy videos from routine practice.


\section{Materials and Methods}

\subsection{Dataset}
\label{sec:dataset}
Colonoscopy videos were collected from patients undergoing routine colonoscopy. An Olympus PCF-H190 colonoscope with a CLV-190 light source and image processor was used. Videos were recorded at $1920\times 1080$ resolution at 30 frames per second (FPS). To train and validate the proposed localization system, the collected colonoscopy videos were divided into two datasets.

\begin{figure}[] 
\includegraphics[width=3.4in]{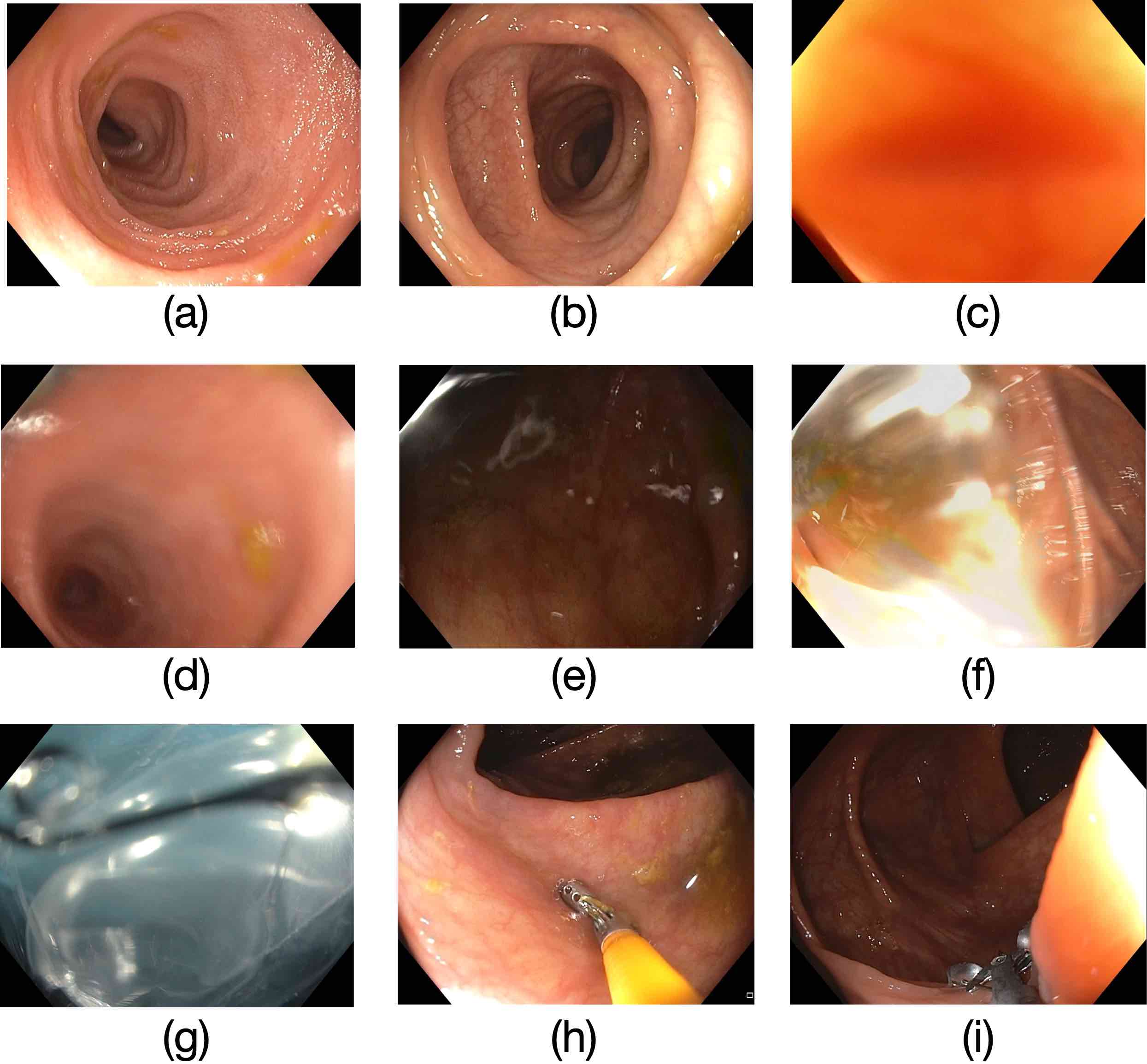}
\caption{Examples of informative frames (a-b); non-informative frames (c-g); and frames with biopsy forceps (h-i). Frames (a) and (b) are examples of informative frames at different colon regions; (c) is a frame captured when the camera was too close to the colon wall; (d) is a frame with significant motion blur; (e) is an underexposed frame; (f) is an overexposed frame; (g) is a frame captured outside of the colon; and (h) and (i) are frames in which biopsies were performed.}
\label{fig:image_examples}
\end{figure}

\textbf{Image classification dataset}: Frames were sampled at 1 FPS from 29 colonoscopy videos. In total, the image classification dataset contains 34,810 frames, which were manually annotated by a gastroenterologist. In this study, frames captured in close proximity to the colon wall, with significant motion blur, with over- or under-exposure, or those captured outside of the body were annotated as non-informative. Figures \ref{fig:image_examples} (a)-(b) present examples of informative frames while Figures \ref{fig:image_examples} (c)-(g) present examples of non-informative frames. 24,425 of 34,810 frames were annotated as non-informative (labeled as ``1''), with the median percentage of non-informative frames in each colonoscopy video being 59.8\%. For biopsy forceps detection, 932 frames were manually annotated as ``with biopsy forceps". The median percentage of frames with biopsy forceps is 3.0\%. Examples of frames with biopsy forceps are shown in Figures \ref{fig:image_examples} (h)-(i). The image classification dataset was randomly split into the training set ($n=19$) and test set ($n=10$). The training set was used for hyper-parameter tuning and model training. The test set was used to evaluate the performance of the trained models.  

\textbf{Localization Dataset}: The localization dataset consists of 44 colonoscopy videos. The localization dataset was divided into three subsets. Sixteen videos were used to build the camera motion estimation network (Set 1); eighteen colonoscopy videos were used for colon template building (Set 2), and the remaining ten videos, which were paired with ScopeGuide videos, were used as an independent evaluation of the localization algorithm (Set 3). Set 1 was further randomly divided into a training set ($n=10$), validation set ($n=3$), and test set ($n=3$). Each video contains over 3,000 samples, and in total, there are 36,665 samples in the training set. The training and validation set were used for hyper-parameter tuning. The test set was used to evaluate the performance of the trained models. For all videos in the localization dataset, the time point at which the camera was withdrawn was manually annotated by the colonoscopy performer. For all videos in the localization dataset, the camera was withdrawn at the cecum.

\textbf{EndoSLAM Dataset} \citep{ozyoruk2020quantitative}: The dataset provides videos from different cameras on ex-vivo porcine gastrointestinal organs. A robotic arm was used to track the camera trajectory and quantify the six degree-of-freedom (DoF) pose values. Six videos from three trajectories (Colon-IV Trajectory-2, Small Intestine-IV Trajectory-1, Stomach-II Trajectory-4) were publicly available with the corresponding ground truth for the camera pose. Each trajectory was recorded by a high-resolution endoscopic camera and a low-resolution endoscopic camera.

\subsection{Image classification}
Our preliminary work on non-informative frame classification was described in \citep{yao2019automated}. In that work, hand-crafted features were combined with bottleneck features for image classification. However, we found that when applying this method to the larger dataset of the current study, the deep learning method alone achieved comparable performance. As a result, in this study, a CNN was used directly for image classification.

Pre-processing was first performed on frames sampled from the colonoscopy videos for standardization. The frames were binarized to identify the largest 4-connected component. After that, the smallest bounding box containing the largest component was used to crop the image. Zero paddings were added to fill the rectangular region into a square region, after which the image was resized to $256\times 256$. These resized images were then inputted into the CNN.

In this study, the Inception-v3 architecture \citep{szegedy2016rethinking} was used for non-informative frame classification. Non-informative frame detection and removal can reduce computational load and avoid unexpected errors in camera localization. Similarly, another deep learning model with Inception-v3 architecture was built to detect frames with biopsy forceps. During a biopsy, the scene between consecutive frames may not be static, but the colonoscope stays at the similar location. In this work, we identified frames during a biopsy by detecting frames with biopsy forceps. After frames with biopsy forceps were detected, frames at 1 second before or after the detected frame were regarded as frames where a biopsy exists. Removing those frames can avoid inaccurate camera motion estimation from the non-rigid scene. The choice of 1 second was made empirically by watching the colonoscopy videos. 
The Inception-v3 architecture is a 42-layer CNN. It was chosen for image classification in this study because it has achieved success in multiple visual tasks \citep{szegedy2016rethinking}. Considering the large number of parameters for Inception-v3, in the training phase the networks were initialized using a pre-trained model on ImageNet \citep{russakovsky2015imagenet}. For biopsy forceps detection, only the last fully-connected layer was fine-tuned using the training set. L2 regularization and dropout were used to improve the model's generalization.

\begin{figure*}[]
\centering
\includegraphics[width=6.6in]{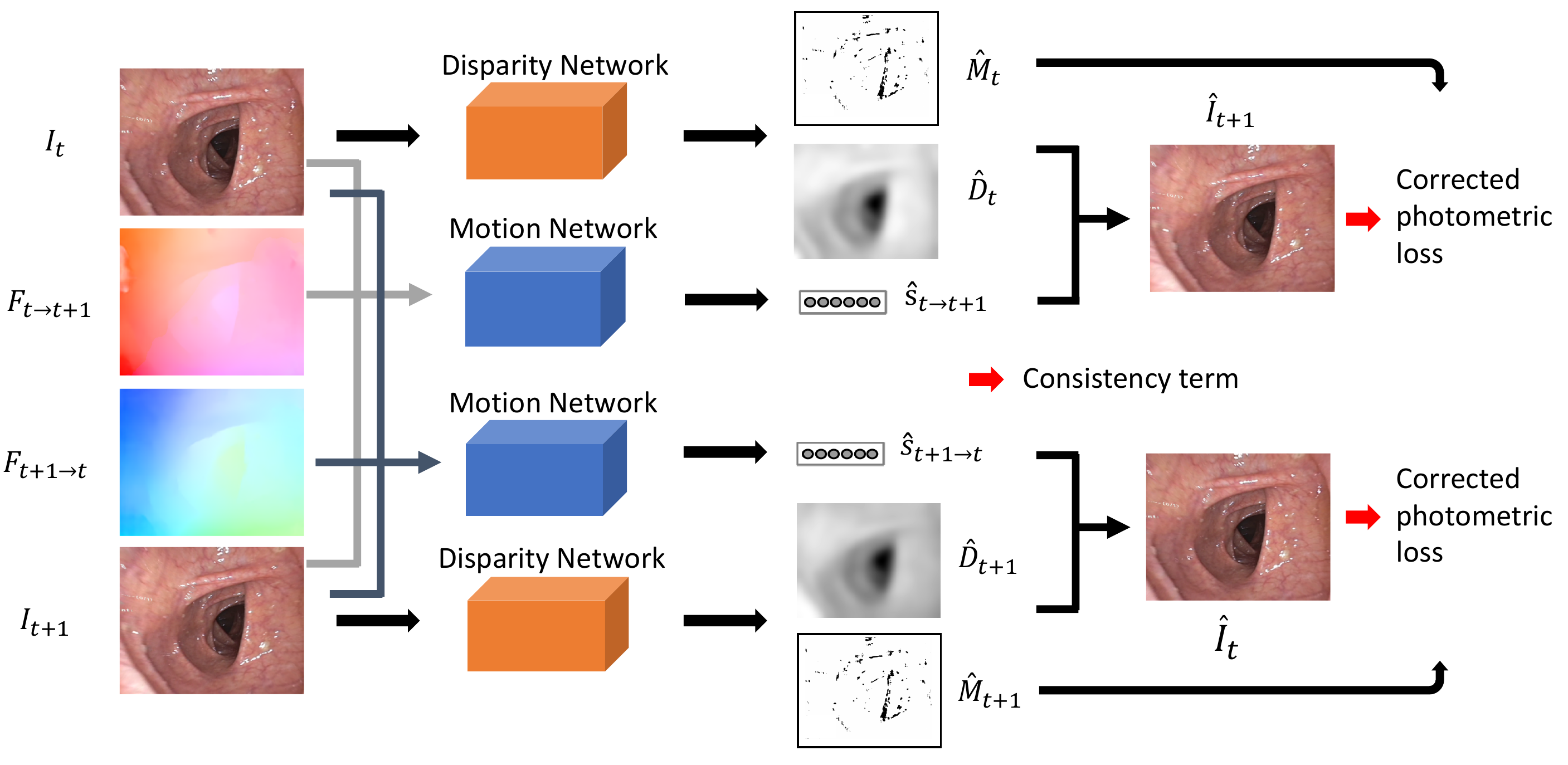}
\caption{Camera motion estimation network. The input to the network is a pair of original frames $I_t,I_{t+1}$ and corresponding optical flows $F_{t\rightarrow t+1},F_{t+1\rightarrow t}$ (visualized using color-coding). The network consists of two sub-networks: the disparity network and the motion network. Loss sources are given following the red arrows.}
\label{fig:network}
\end{figure*}

\subsection{Camera motion estimation}
\subsubsection{Preprocessing}
In this study, the camera at the tip of the colonoscope has a fisheye lens (Olympus PCF-H190). Camera calibration was performed to estimate the camera's intrinsic matrix. With the mathematical model of a fisheye camera proposed in \citep{scaramuzza2006toolbox}, the distorted images from colonoscopy videos were corrected. Details of the camera model and image distortion correction are covered in \ref{sec:camera_model}.

\subsubsection{Architecture}
After the camera calibration and image distortion correction, the corrected frames were used for camera motion estimation. An overview of the camera motion estimation architecture is depicted in Figure \ref{fig:network}. Let us denote a corrected frame from a colonoscopy video with size $H\times W\times C$ at time point $t$ as $I_t: \Omega_{I}\to [0, 1]$, where $\Omega_{I}=\{1,2,\dots,H\}\times \{1,2,\dots,W\}\times\{1,2,\dots,C\}$. The unit of time is arbitrary, with a smaller time duration from $t$ to $t+1$ allowing for better estimation. In this study, the duration from $t$ to $t+1$ was chosen to be $\frac{1}{15}$ second. Given a pair of consecutive frames $I_t$, $I_{t+1}$, the dense optical flow can be calculated using PWC-Net \citep{sun2018pwc} (the details of which are discussed in \ref{ap:optical_flow}), which achieved the highest accuracy on several published datasets. Optical flow is a way to describe the magnitude and orientation of apparent velocities of brightness within an image. The dense optical flow from $I_t$ to $I_{t+1}$ is denoted as $F_{t\to t+1}: \Omega_{F}\to \mathbb R$, where $\Omega_{F}=\{1,2,\dots,H\}\times \{1,2,\dots,W\}\times\{1,2\}$. The pattern in optical flow $F_{t\to t+1}$ shows how a rigid scene changes with the camera's motion from $t$ to $t+1$. After the optical flow calculation, a concatenation of $I_t$, $I_{t+1}$, $F_{t+1\to t}$ is fed into the motion network to estimate the 6 DoF of the camera's motion: $\hat s_{t\to t+1}=[\hat t_x, \hat t_y, \hat t_z, \hat r_x,  \hat r_y, \hat r_z] \in \mathbb R ^6$. Simultaneously, $I_t$ is fed into the disparity network to estimate the corresponding disparity map $\hat D_{t}:  \Omega_{D}\to [0, 1]$, where $\Omega_{D}=\{1,2,\dots,H\}\times \{1,2,\dots,W\}$, and the specular region mask $\hat P_{t}:  \Omega_{P}\to [0, 1]$, where $\Omega_{P}=\{1,2,\dots,H\}\times \{1,2,\dots,W\}$. In this study, a value in the disparity map is defined as the inverse of the corresponding scene depth. Similarly, the optical flow $F_{t+1\rightarrow t}$ from $I_{t+1}$ to $I_t$ is calculated, and $\hat s_{t+1 \to t}$, $\hat D_{t+1}$, $\hat P_{t+1}$ are estimated by the motion network and disparity network, respectively. In this study, the estimated disparity map was only used to facilitate the loss calculation. After model training, only the motion network was used for camera localization.

Figure \ref{fig:a.network_architecture_motion} in \ref{sec:a.architecture} shows the detailed structure of the motion network. In the motion network, the input is either \{$I_t$; $I_{t+1}$; $F_{t\to t+1}$\} or \{$I_{t+1}$; $I_{t}$; $F_{t+1\to t}$\}. While the optical flow pattern can provide information about the camera's motion, the original frames can provide more information about the scene structure, reducing ambiguity in motion detection. The motion network contains eight convolutional layers. The first seven convolutional layers have a filter size of $3\times 3$ and are followed by a max-pooling layer and a ReLU activation function. The last convolutional layer has six filters of size $1\times 1$. The first three filters in the last convolutional layer are used to estimate the predicted camera translation in the $x$-, $y$-, and $z$-axes, while the last three filters are used to estimate Euler angles of the predicted camera rotation. The last convolutional layer is followed by a global average pooling layer to aggregate predictions at all spatial locations. Using the max-pooling layers the motion network can capture both global and local optical flow patterns for camera motion estimation. 

Figure \ref{fig:a.network_architecture_disparity} in \ref{sec:a.architecture} shows the detailed structure of the disparity network. The input of the disparity network is a corrected frame. Previous literature has shown that non-linear transformations can be modeled to convert a single view image to its corresponding disparity map \citep{garg2016unsupervised, mayer2016large}. The disparity network has an encoder-decoder structure. Unlike in previous literature, the disparity network performs two tasks: disparity estimation and specular region estimation. Considering that both tasks involve intensity and low-level textural feature analysis, a multi-task strategy is applied here, where the two tasks share the same encoder and then have two individual decoders with similar architectures. The encoder consists of three convolutional layers and two max-pooling layers, while the decoder consists of three convolutional layers and two up-sampling layers. All convolutional layers have a filter size of $3\times 3$. For disparity estimation, the last convolutional layer is followed by a sigmoid activation function and a multiplication with 10 to constrain every entry in the estimated disparity map to $[0, 10]$. For specular region estimation, the last convolutional layer is followed by a softmax activation function. All other convolutional layers are followed by a ReLU activation function. The encoder-decoder structure facilitates local and global information extraction and integration.

\subsubsection{Loss function}
\noindent
\textit{(A) Regular photometric loss}

For camera motion estimation, the photometric error between the synthesized new frame and the real frame is used as the loss function. Given a frame $I_t$, estimated disparity map $\hat D_{t}$, and estimated camera motion $\hat s_{t \to t+1}$, a new frame is synthesized at time point $t+1$ by applying differentiable image warping. 

Let $p_t$ denote the coordinate of one pixel in the image plane at time point $t$. If we assume the world frame and the camera frame at time point $t$ are the same, from equations (\ref{equ:world-to-camera}) and (\ref{equ:camera-to-image}), the homogeneous pixel coordinates can be projected back to the 3D world coordinate as:
\begin{equation}
    p_w = \hat D_t(p_t)^{-1}K^{-1}p_t,
\label{equ: world_coords}
\end{equation}
where $p_t = [x, y, 1]\tran$, $x\in [1,2,\dots, H], y\in [1,2,\dots, W]$, are the homogeneous pixel coordinates in $I_t$, and $p_w = [X, Y, Z]\tran, X \in \mathbb{R}, Y\in \mathbb{R}$, and $Z\in \mathbb{R}$, are the 3D world coordinates of the object shown at $p_t$. 

After the camera's motion $\hat s_{t \to t+1}$, $p_t$ will be projected onto the new image plane as:
\begin{subequations}
\begin{align} 
    &p_{t+1} = K(\hat Rp_w+\hat T) \\
    &\hat R = R_x(\hat r_x)R_y(\hat r_y)R_z(\hat r_z)\\
    &R_x(\hat r_x) = \begin{pmatrix}
1&0&0\\
0&\cos \hat r_x& -sin \hat r_x\\
0&sin \hat r_x& \cos \hat r_x
\end{pmatrix}\\
    &R_y(\hat r_y) = \begin{pmatrix}
\cos \hat r_y&0& \sin \hat r_y\\
0&1&0\\
-\sin \hat r_y&0&\cos \hat r_y
\end{pmatrix}\\
    &R_z(\hat r_z) =\begin{pmatrix}
\cos \hat r_z &-\sin \hat r_z&0\\
\sin \hat r_z&\cos \hat r_z &0\\
0&0&1
\end{pmatrix}\\
&\hat T = \begin{pmatrix}
\hat t_x\\
\hat t_y\\
\hat t_z
\end{pmatrix}
\end{align}
\label{equ: new_image_coords}
\end{subequations}

\begin{figure}[] 
\centering
\includegraphics[width=3.2in]{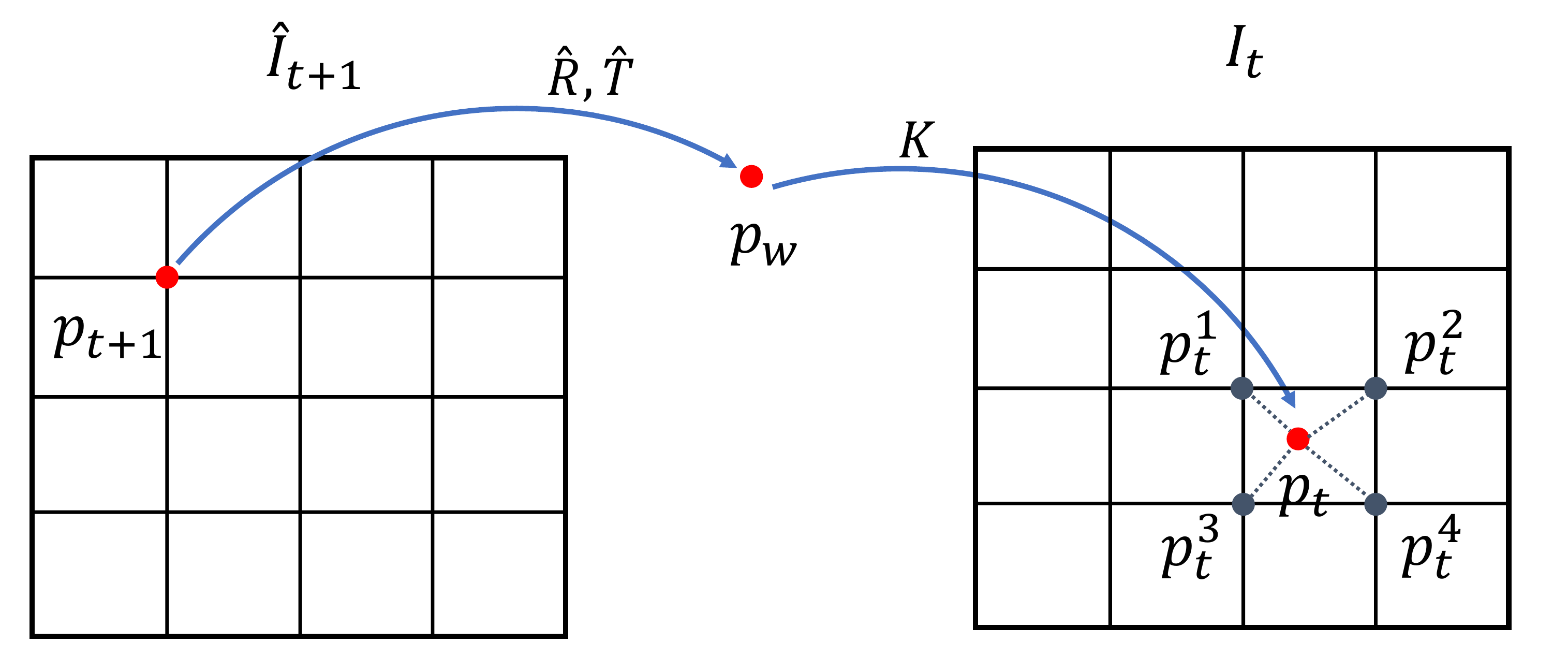}
\caption{A diagram of the image warping process. To synthesize $\hat I_{t+1}$, the coordinate $p_{t+1}$ in $\hat I_{t+1}$ can be projected back to the world coordinate frame as $p_w$, from which $p_{t}$ in $I_{t}$ can be calculated. $p_w$ is the world coordinate of an object, and $p_t$, $p_{t+1}$ are the coordinates of this object shown in $I_t$, $I_{t+1}$, respectively. We assume that the intensity value $I_{t+1}(p_{t+1})$ is equal to $I_t(p_t)$. The intensity value at $p_t$ can be approximated using bilinear interpolation with its four neighbors.}
\label{fig:image_warp}
\end{figure}

From equations (\ref{equ: world_coords}) and (\ref{equ: new_image_coords}), all pixel coordinates in $I_t$ can be projected back into the world coordinates, and then projected to the estimated image plane $I_{t+1}$, and vice versa. An image warping process is used to generate the new frame $\hat I_{t+1}$.  Figure \ref{fig:image_warp} presents a diagram of the image warping process. Given a pixel coordinate $p_{t+1}$ from $
\hat I_{t+1}$, $p_t$, the corresponding coordinates of this object at $I_t$ can be calculated using equations (\ref{equ: world_coords}) and (\ref{equ: new_image_coords}). The intensity value $I_{t+1}(p_{t+1})$ is the same as $I_{t}(p_{t})$ based on two assumptions: that the colon surface exhibits Lambertian reflectance and the camera motion between consecutive frames is very small. As shown in Figure \ref{fig:image_warp}, to calculate $\hat I_{t+1}(p_{t+1})$, the estimated intensity at the position $p_{t+1}$ in the new frame, bilinear interpolation is used to approximate $I_t(p_t)$ using its four pixel neighbors $p_t^1$, $p_t^2$, $p_t^3$, $p_t^4$ as
\begin{equation}
    \hat I_{t+1}(p_{t+1}) = I_t(p_t) = \sum_{i=1}^{4} w_iI_t(p_t^i),
\end{equation}
where $w_i$ is the relative spatial distance between $p_t$ and $p_t^i$, and
\begin{equation}
\sum_{i=1}^{4} w_i = 1.
\end{equation}
We denote the warping process from $I_t$ to $\hat I_{t+1}$ as $\mathcal{W}_{t \to t+1}$, thus we have $\mathcal{W}_{t \to t+1}(I_t)=\hat I_{t+1}$

With the image warping process, $\hat I_{t+1}$ is synthesized. Similarly, $\hat I_{t}$ can be synthesized given $\hat s_{t+1\to t}$, $I_{t+1}$, and $\hat D_{t+1}$. The photometric loss for the pair $I_{t}$, $I_{t+1}$ can be calculated as

\begin{equation}
loss_{p} = \frac{1}{Z}\sum_{p \in \Omega_I} \|I_t(p)-\hat I_t(p)\|^2 +
    \frac{1}{Z}\sum_{p \in \Omega_I} \|I_{t+1}(p)-\hat I_{t+1}(p)\|^2,
\end{equation}
where $Z=H\cdot W\cdot C$.

\bigskip
\noindent
\textit{(B) Corrected photometric loss}

During the calculation of $\hat I_{t+1}$, the calculated $p_t$ may be out of the frame $I_t$, and usually, a value of 0 will be assigned. The photometric differences in those regions should not be used as supervision for motion estimation because they result from missing information. In this study, an intensity of $-1$ was assigned instead. A mask $M_{t+1}$ can then be built for $\hat I_{t+1}$, where $M_{t+1}(p)=1$ if $\hat I_{t+1}(p)=-1$ otherwise $M_{t+1}(p)=0$. The mask $M_{t+1}$ can be used to filter out invalid pixel locations that were mapped out of the frame $I_t$.

\begin{figure*}[] 
\centering
\includegraphics[width=6.8in]{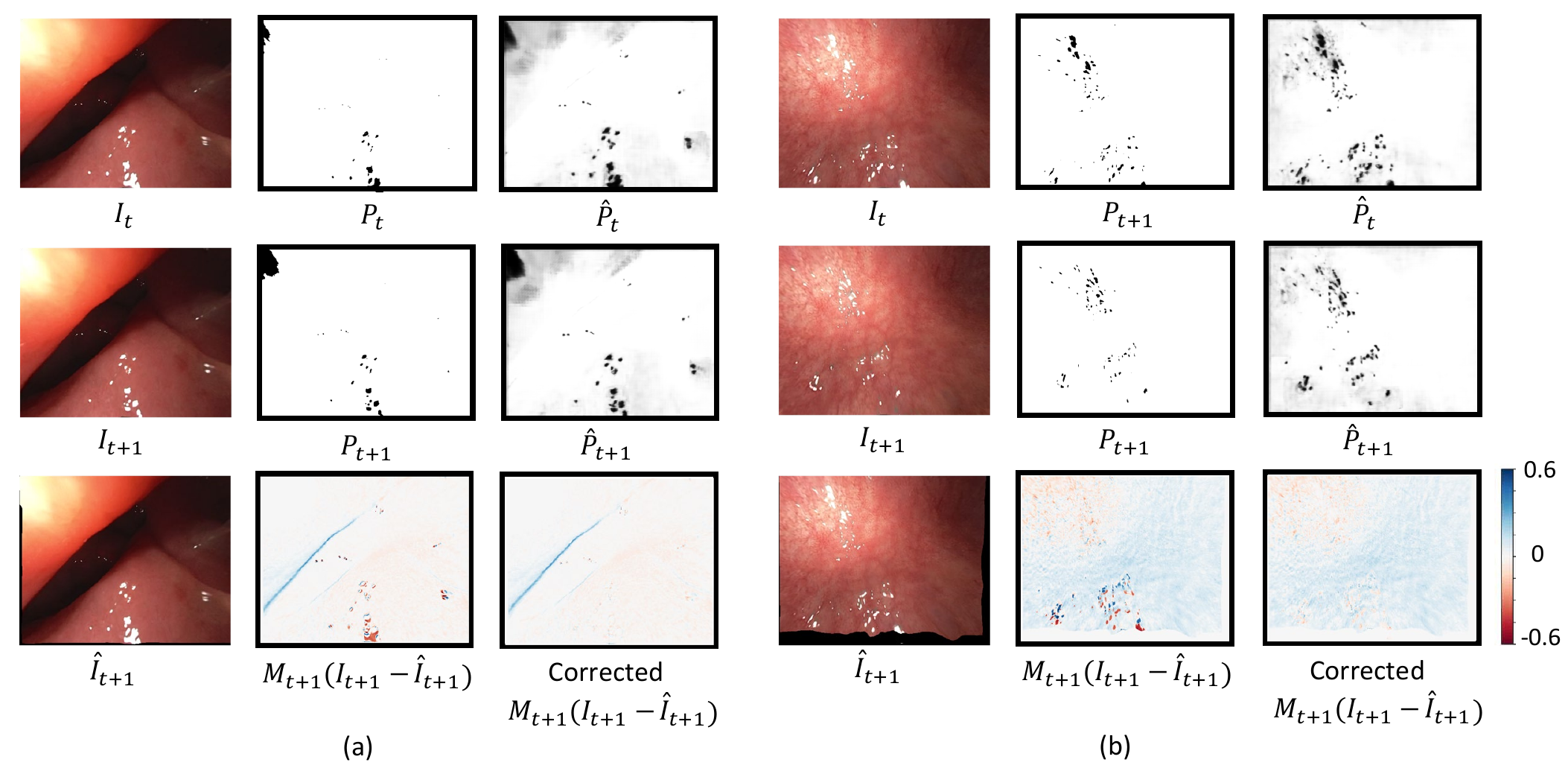}
\caption{Two example pairs of consecutive frames on specular mask estimation. Threshold-based specular mask (middle column) and the specular mask estimated from the disparity network (right column) are presented. The photometric error map between the projected image $\hat I_{t+1}$ and $I_{t+1}$ without and with the estimated specular mask are shown.}
\label{fig:specular_mask}
\end{figure*}

In the image warping process, we assume the colon surface exhibits Lambertian reflectance, which is not true for specular regions. Figure \ref{fig:specular_mask} shows two examples in which the brightness of specular regions within $I_t$ and $I_{t+1}$ changes significantly with different angles of view. As such, pixel locations in specular regions should be excluded when the photometric loss is calculated. In our previous work, we attempted to calculate specular region masks by converting the RGB image into HSV space and then extracting the specular region using a threshold on the saturation channel. Figure \ref{fig:specular_mask} shows examples of threshold-based mask as $P_{t}$ and $P_{t+1}$. While this method is simple and can provide information on the specular region, it is very hard to find an optimal threshold that generalizes for all frames and videos. To estimate the specular region more accurately, one should also consider intensity statistics and textural features. As such, a branch of the disparity network was used for specular region estimation. As shown in Figure \ref{fig:a.network_architecture_disparity}, another decoder is used to estimate the specular region mask, and a softmax activation function is applied to the feature map from the last layer to generate a probabilistic map. Examples of the output $\hat P_{t}$, $\hat P_{t+1}$ are shown in Figure \ref{fig:specular_mask}, where a smaller value indicates the location is more likely to be within a specular region. With estimated $\hat s_{t\to t+1}$, $\hat D_{t+1}$, the same image warping process $\mathcal{W}_{t\to t+1}$ will be applied to $\hat P_{t}$ to generate the specular region mask for $\hat I_{t+1}$. 

Based on the estimated masks  $\hat P_{t}$, $\hat P_{t+1}$, and $M_{t}$, $M_{t+1}$, a corrected photometric loss can be written as:

\begin{equation}
\small
\begin{split}
    &loss_{cp} = \frac{1}{Z}\sum_{p \in \Omega_I} M_t(p)*\|\hat P_{t}(p)*I_t(p)- \mathcal{W}_{t+1 \to t}(\hat P_{t+1})(p)*\hat I_t(p)\|^2 +\\
    &\indent \frac{1}{Z}\sum_{p \in \Omega_I} M_{t+1}(p)*\|\hat P_{t+1}(p)I_{t+1}(p)-\mathcal{W}_{t \to t+1}(\hat P_{t})(p)*\hat I_{t+1}(p)\|^2,
\end{split}
\end{equation}
where $*$ denotes element-wise multiplication.

An additional cross-entropy loss was calculated to train the model for specular region estimation:
\begin{equation}
loss_{ce} = cross\_entropy(P_t, \hat P_t) + cross\_entropy(P_{t+1}, \hat P_{t+1}), 
\end{equation}
where $P_t$ is calculated using a threshold of 0.1 determined by visual evaluation. $loss_{ce}$ is proposed as a weak supervision for specular region estimation.

A combination of $loss_{ce}$ and $loss_{cp}$ encourages the network to detect the specular region and also minimize the photometric error. Figure \ref{fig:specular_mask} shows examples of $\hat P_t$ and $\hat P_{t+1}$. The disparity network detects more comprehensive specular regions as compared to the threshold-based method.

The last row of Figure \ref{fig:specular_mask} depicts the benefits of calculating $loss_{cp}$. The movements of the edge segment in (a) and vessel pattern in (b) should be the primary cues used to estimate the camera's motion. However, the photometric difference between $I_{t+1}$ and $\hat I_{t+1}$ around the specular region is very high, which may overwhelm the photometric difference from other regions. As a result, with $loss_p$, the network will be encouraged to reduce the photometric difference on specular regions with a higher priority. Considering the specular regions are not Lambertian surfaces, the photometric differecne on specular regions may be inaccurate. The problem can be fixed by applying the estimated specular mask to correct the photometric loss.

\bigskip
\noindent
\textit{(C) Movement consistency loss}

The forward movement $\hat s_{t\to t+1}$ and backward movement $\hat s_{t+1 \to t}$ are estimated by the motion network. Let us denote the transformation matrix from frame $t$ to $t+1$ as $Q_{t\to t+1} \in \mathbb{SE}(3)$ and denote its inverse as $Q_{t+1\to t}$, then

\begin{equation}
Q_{t\to t+1} \times Q_{t+1\to t} =  Q_{t\to t}=I,
\label{eq:consistency}
\end{equation}
where $\times$ denotes matrix multiplication and $I$ denotes the identity matrix.

As there is no camera movement from frame $t$ to itself, $Q_{t\to t}$ is an identity transformation in homogeneous coordinates.

A movement consistency loss term can be written as 
\begin{subequations}
\begin{align} 
    &loss_{mc} = \|\hat Q_{t \to t+1} - \hat Q_{t+1 \to t}^{-1}\|_F,\\
    &\hat Q_{t \to t+1} = \begin{pmatrix}
    \hat R_{t \to t+1}&\hat T_{t \to t+1}\\
    \mathbf{0}_{1\times 3}&1
    \end{pmatrix},\\
    &\hat Q_{t+1 \to t} = \begin{pmatrix}
    \hat R_{t+1 \to t}&\hat T_{t+1 \to t}\\
    \mathbf{0}_{1\times 3}&1
    \end{pmatrix},\\
    &\hat Q_{t+1 \to t}^{-1} = \begin{pmatrix}
    \hat R_{t+1 \to t}^T&-\hat R_{t+1 \to t}^T\times T_{t+1 \to t}\\
    \mathbf{0}_{1\times 3}&1
    \end{pmatrix}.
\end{align}
\label{equ: consistency_loss}
\end{subequations}
 
The movement consistency term encourages the network to output a forward movement $\hat s_{t\to t+1}$ and backward movement $\hat s_{t+1 \to t}$ that satisfy equation (\ref{eq:consistency}).The movement consistency term was added into the final loss to improve the network's generalization.

\bigskip
\noindent
\textit{(D) Final loss}

The final loss function can be written as:
\begin{equation}
\label{eq:final_loss}
loss = loss_{cp} + \lambda_{ce}loss_{ce} + \lambda_{mc}loss_{mc} + \lambda_{smo}loss_{smo},
\end{equation}
where $loss_{smo}$ is from previous literature \citep{wang2018learning} on monocular depth estimation and is used to encourage the estimated disparity map to be locally smooth. $\lambda_{ce}$, $\lambda_{mc}$, and $\lambda_{smo}$ are loss weights.

\begin{figure}[] 
\centering
\includegraphics[width=2.8in]{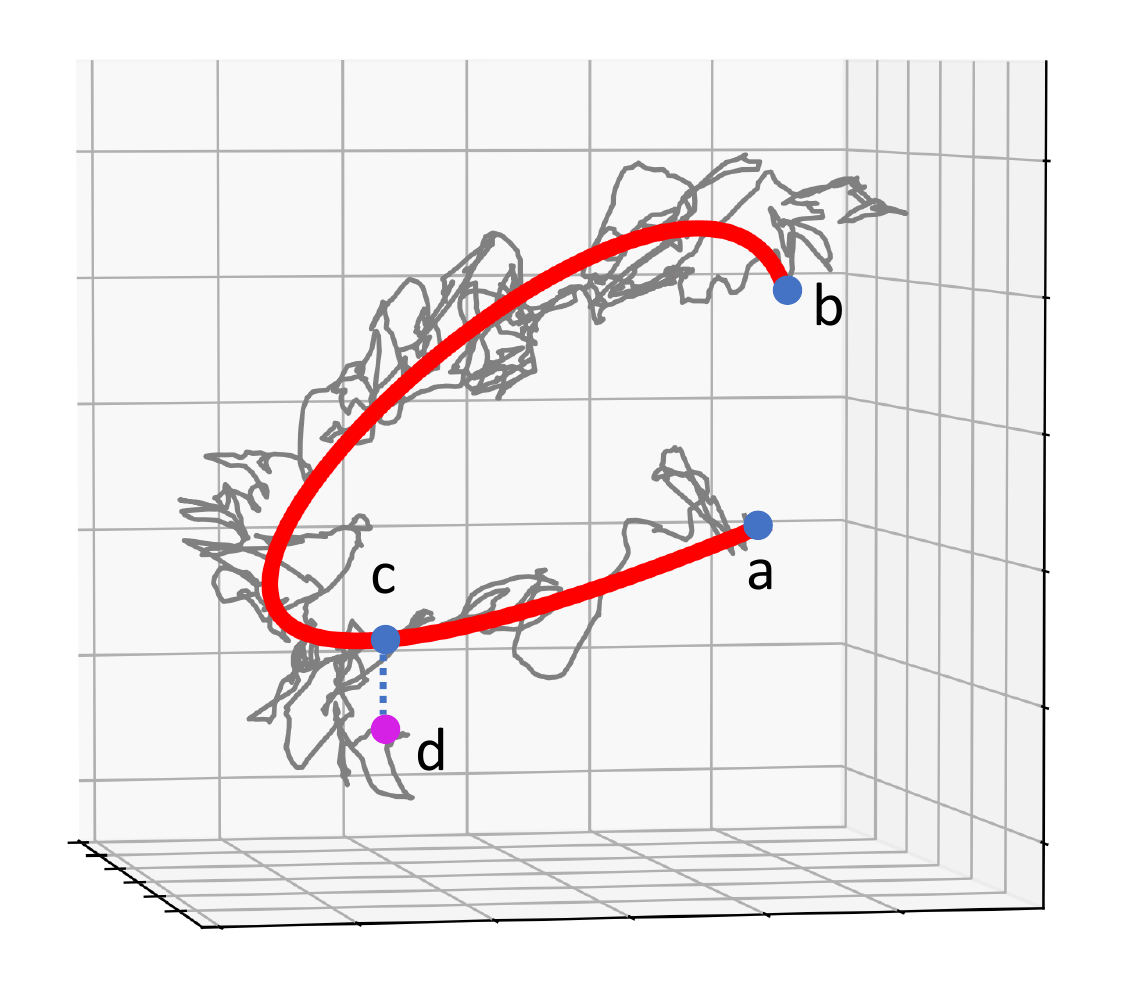}
\caption{A diagram of the location index estimation. The gray line is a camera trajectory and the red line is the major traveling path. $a$, $b$ are the start and end coordinates in the major traveling path of the camera, respectively. $d$ is the coordinate of the frame at the time point $t$ in the camera trajectory, and $c$ is the closest point on the red line to $d$.}
\label{fig:location_index}
\end{figure}

\subsection{Camera trajectory and location index estimation}
The camera motion network was applied to colonoscopy video in the withdrawal phase to enable successive estimation of the camera's motion, along with the coordinates of the camera (i.e., camera trajectory) after the camera's withdrawal begins. The relative location index is defined from 0 to 1. When the camera starts to be withdrawn at the beginning of the colon (cecum), the frame's location index is 0; when the camera stops at the end of the colon (rectum), the frame's location index is 1. Let $l_t$ denote the colon length that the camera traversed from the beginning to time point $t$ and $l_{all}$ denote the total length of the colon from the cecum to rectum. The location index for the frame at time point $t$ can be calculated as $l_t/l_{all}$.

The length of the camera trajectory (gray line in Figure \ref{fig:location_index}) can not be used directly to estimate the location index as the camera may be moved about to inspect the colon's surface. Considering that the shape of the colon is curved, we propose to estimate a major traveling path for location index calculation. Figure \ref{fig:location_index} depicts how the location index is calculated. The gray line is a camera trajectory in the withdrawal phase. The camera trajectory zigzags because the camera will turn about, moving back and forth to inspect the colon mucosal surface. The red line is the major traveling path of the camera. To estimate the major traveling path of the camera, we used a B-spline curve fitting algorithm with a smoothing factor of $\gamma$ to fit the original camera trajectory. Let us assume $a$ is the beginning of the colon; $b$ is the end of the colon; and $d$ is the position of the camera at time point $t$. $l_t$ can be calculated as the length of the red line from $a$ to $c$, and $l_{all}$ can be calculated as the length of the red line from $a$ to $b$. Then the location index for the frame at the time point $t$ is the ratio of the length from $a$ to $c$ over the length from $a$ to $b$ in the red line.

The calculation of the frame's location index can convert the frame from the time domain to the location domain. A frame with a location index $x$ can be denoted as $I_x^d$. Let us denote $f_v:[1,2,\dots, N]\to [0, 1]$ as a function that maps the time index of a frame to a location index in the colonoscopy video $v$, with $N$ being the maximal time index. Given $I_t$ - a frame at the time point $t$ in $v$ - the estimated location index for this frame can be written as $f_v(t)$, and the frame can then be denoted as $I_{f_v(t)}^d$. Figures \ref{fig:template}(a)-(c) show examples of time-location mapping. If the time domain is used, the frames can be denoted as $I_1, \dots, I_t, \dots, I_N$, where $1\leq t\leq N$; if the location domain is used, the frames can be denoted as $I_{f_v(1)}^d, \dots, I_{f_v(t)}^d, \dots,I_{f_v(N)}^d$. If $I_a$ and $I_b$ are the frames at the beginning and end of the colon, respectively, we have $f_v(a)=0$ and $f_v(b)=1$.

\subsection{Anatomical colon segment classification}

\begin{figure*}[t] 
\centering
\includegraphics[width=5.8in]{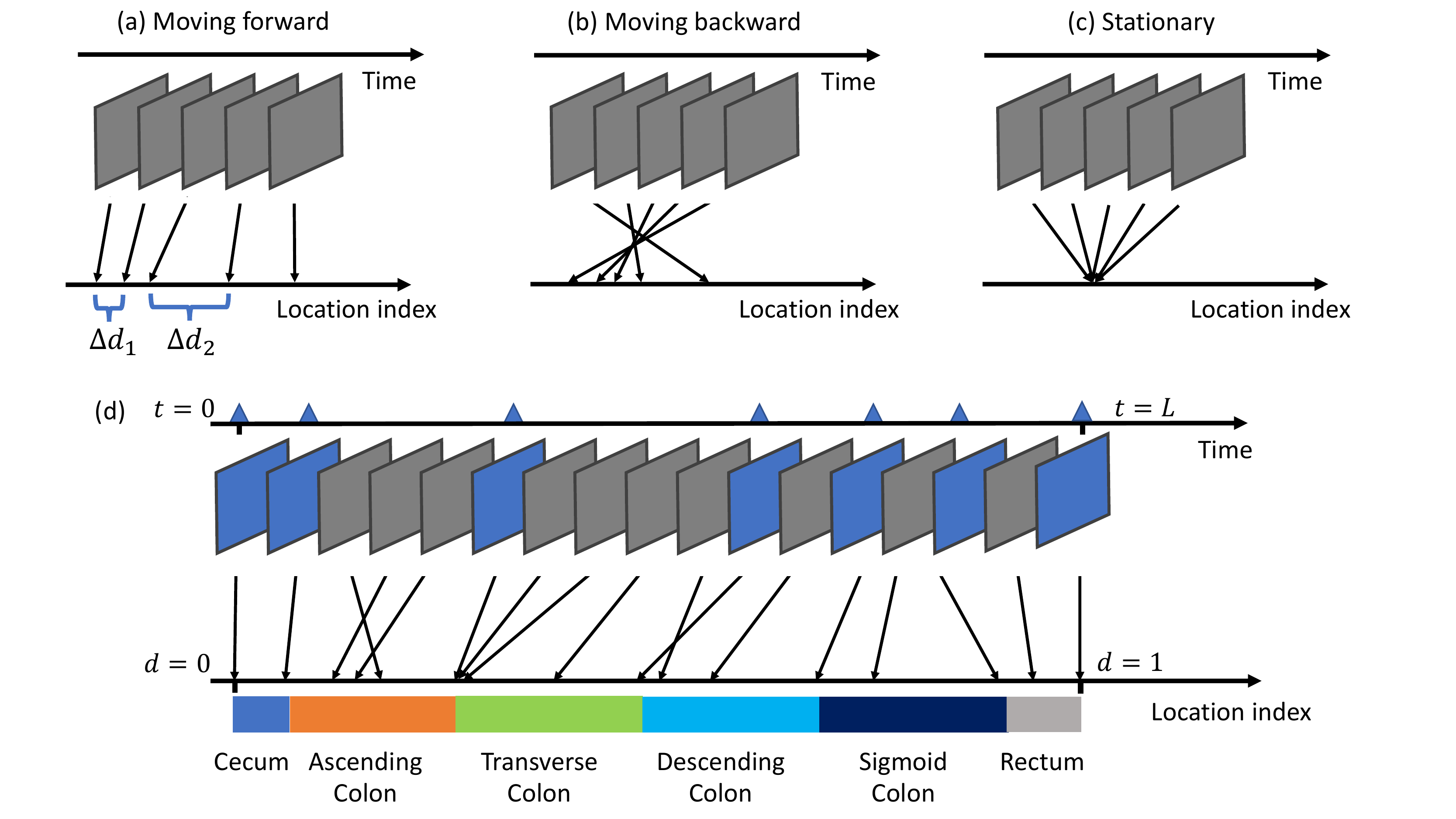}
\caption{A diagram of time-location mapping and colon template building. (a)-(c) are examples when the camera moves forward, backward, or is stationary. $\Delta d_1$ and $\Delta d_2$ are the same duration in time, but $\Delta d_2$ is much larger than $\Delta d_1$. (d) shows a full colonoscopy frame sequence in the withdrawal phase mapped into the location domain. To build the template, time points when the colonoscope enters and exits each colon segment were annotated by a physician (blue triangles). Based on the time annotations and time-location mapping, the relative length of each colon segment can be calculated.}
\label{fig:template}
\end{figure*}

Anatomical colon segment classification can be performed based on the calculated colon location index. It not only provides contextual information for severity assessment but also helps to validate the performance of the location index estimation. In the proposed anatomical colon segment classification method, we assume that the relative length of each colon segment is similar across patients. The colon segments used in this study include cecum, ascending colon, transverse colon, descending colon, sigmoid colon, and rectum. In anatomical colon segment classification, we assume that camera withdrawal begins at the cecum.  

To build the colon template, the the times at which the camera enters each colon segment in the withdrawal phase were annotated. As shown in Figure \ref{fig:template}(d), given a colonoscopy video $v$, a vector $q_{v}^t$ contains 7 times manually annotated by physicians as
\begin{equation}
q_{v}^t = [t_v^{1}, t_v^{2}, t_v^{3}, t_v^{4}, t_v^{5}, t_v^{6}, t_v^{7}],
\end{equation}
where the first 6 entries are the times at which the camera enters the rectum, ascending colon, transverse colon, descending colon, sigmoid colon, and cecum in the withdrawal phase, respectively, and the last entry is the time point when the camera stops. A corresponding vector $q_{v}^s$ containing the location index of the frames when the camera enters each colon segment and stops at the rectum can be estimated as
\begin{equation}
q_{v}^s[i] = f_v(q_{v}^t[i]),
\end{equation}
where $q_{v}^s[i]$ is the $i^{th}$ element of $q_{v}^s$, $1\leq i\leq7$, and $f_v(q_{v}^t[1])=0$ and $f_v(q_{v}^t[7])=1$. 

The relative length of each colon segment in the colonoscopy video $v$ can be recorded in the vector $q_v^{rl}$ as
\begin{equation}
q_{v}^{rl}[i] = f_v(q_{v}^t[i+1])-f_v(q_{v}^t[i]),
\end{equation}
where $q_{v}^{rl}[i]$ is the $i^{th}$ element of $q_{v}^{rl}$, and $1\leq i\leq6$.

The colon template can be estimated by annotating the times at which the camera enters each colon segment for $N$ colono-scopy videos $v_1, v_2, \dots, v_N$. The colon template $q^{ct}$ can be estimated as
\begin{equation}
q^{ct}[i] = \frac{w}{N}\sum_{j=1}^{N}{q_{v_j}^{rl}[i]},
\end{equation}
where $1\leq i\leq6$, and $w$ is a scaling factor and
\begin{equation}
\sum_{i=1}^{6}{\frac{w}{N}\sum_{j=1}^{N}{q_{v_j}^{rl}[i]} }=1.
\end{equation}

With a new colonoscopy video, the start of the camera's withdrawal can be identified using the physician's notes, while the end of the camera's withdrawal can be identified as the time of the last informative frame. Location index estimation is then performed for a frame sequence during the withdrawal phase. By comparing these with the constructed colon template, frames falling within each colon segment can be classified.

\subsection{Model training and hyper-parameter tuning}
The image classification and motion estimation models were implemented in Tensorflow v1.10 and were trained on an NVIDIA Tesla V100. Adaptive moment estimation (ADAM) was used for optimization.

For the image classification model, the hyper-parameters, including learning rate, batch size, dropout rate and L2 regularization, were chosen via 5-fold cross-validation on the image classification training set. Based on the classification AUCPR (area under the precision-recall curve), a learning rate of 10$^{-3}$, a batch size of 8, a dropout rate of 0.4, and a L2 regularization of 0.0001 were selected to train the network on the whole training set. The trained model was then tested on the test set. 

For the motion estimation network, Set 1 of the localization dataset was used to build the camera motion estimation model. As mention in section \ref{sec:dataset}, Set 1 was further split into the training set ($n=10$), validation set ($n=3$), and test set ($n=3$). Different combinations of hyper-parameters, including learning rate, batch size, training steps, $\lambda_{ce}$, $\lambda_{mc}$, and $\lambda_{smo}$ were used to train the network on the training set, with the hyper-parameters that achieved the best corrected photometric loss (as discussed in Section \ref{sec:evaluation_strategy}) on the validation set chosen as optimal. Based on the results from hyper-parameter tuning, the final camera motion estimation network was trained with a learning rate of 0.0001, a batch size of 1, 300,000 training steps, a $\lambda_{ce}$ of 0.01, a $\lambda_{mc}$ of 100, and a $\lambda_{smo}$ of 0.02. After hyper-parameter tuning, the trained model was tested on the test set. 

For location index estimation, the smoothing factor $\gamma$ was tuned using anatomical colon segment classification on Set 2 of the localization dataset. A 5-fold cross-validation was performed and $\gamma=100$ achieved the best average classification accuracy.

\subsection{Evaluation strategy}
\label{sec:evaluation_strategy}
\noindent
\textit{(A) Image classification}

The performance of the non-informative frame classification and biopsy detection models were evaluated using AUCPR, area under the receiver operating characteristic curve (AUC), sensitivity, specificity, precision, and accuracy. 

\bigskip
\noindent
\textit{(B) Camera motion estimation}

To evaluate the performance of the camera motion estimation model, a corrected photometric error (CPE) was calculated as:

\begin{equation}
\small
\begin{split}
    &CPE = \frac{1}{Z}\sum_{p \in \Omega_I} M_t(p)*\|P_{t}(p)*I_t(p)- \mathcal{W}_{t+1 \to t}(P_{t+1})(p)*\hat I_t(p)\|^2 +\\
    &\indent \frac{1}{Z}\sum_{p \in \Omega_I} M_{t+1}(p)*\|P_{t+1}(p)I_{t+1}(p)-\mathcal{W}_{t \to t+1}( P_{t})(p)*\hat I_{t+1}(p)\|^2,
\end{split}
\end{equation} for each pair of frames. The average CPE throughout a colono-scopy video was then calculated, with a lower average CPE indicating that the model has a better capacity for camera motion estimation. Please note that the threshold-based specular mask was applied in CPE calculation for fairness.

For the EndoSLAM dataset, the ground truth of the camera's pose was provided. Absolute trajectory error (ATE), relative pose error on translation (RPE Trans.), and rotation (RPE Rot.) can be calculated \citep{prokhorov2019measuring}. ATE measures the distance between the estimated trajectory and the ground truth trajectory. RPE measures the local accuracy of the trajectory over a pair of consecutive frames. The calculation of those metrics can be found in \citep{prokhorov2019measuring}. As the camera motion was estimated with an arbitrary scale, the two trajectories should first be aligned by finding a similarity transformation $S$\citep{umeyama1991least}. Lower ATE, RPE Trans., and RPE Rot. indicate better motion estimation.

\bigskip
\noindent
\textit{(C) Location index estimation}

The evaluation of the location index estimation is challenging as there is no  ground truth for the location index. In lieu of this we propose two ways to evaluate the location index estimation. 

The first method is to compare the anatomical colon segment classification with baselines. Two baselines were considered for anatomical colon segment classification. For the first baseline, the time index is used to build the colon template and the subsequent classification, which assumes that during the withdrawal phase, the proportional amount of time the camera spends within each colon segment is the same for different patients. For a colonoscopy video $v$, a vector $q_v^{rt}$ contains the relative time in each colon segment and can be calculated as
\begin{equation}
    q_v^{rt}[i] = \frac{1}{t_{v}^7-t_{v}^1}(t_{v}^{i+1}-t_{v}^i),
\end{equation}
where $q_v^{rt}[i]$ is the $i^{th}$ element of $q_v^{rt}$ and $1\leq i\leq6$. A colon template can then be built for anatomical colon segment classification. The assumption here should not hold as the velocity of the camera's movement varies and is affected by individual colon segment disease severity and condition. As a result, an accurate location index estimation method should lead to higher accuracy in anatomical colon segment classification than using the time index directly. The second baseline for estimating position used the approximate colonoscope insertion length measured by ScopeGuide.  The limitation of using ScopeGuide length is that although the length of the inserted tube provides an approximate location index, it is a measure of the amount of scope inserted rather than the true distance the camera traveled. An introduction and discussion of ScopeGuide length can be found in \ref{appendix:scopeguide}.

To evaluate the classification performance, we denote the colon segments from cecum to rectum as 1-6. The multi-class classification accuracy, the maximal absolute difference between the predicted class and annotated class, and the average absolute difference between the predicted class and annotated class were calculated for each video. The averaged value and standard deviation among videos were then calculated. For each colonoscopy video in the test set, an individual confusion matrix was calculated, with the entry ($i, j$) being the percentage of frames in colon segment $i$ classified as in colon segment $j$, where $i, j \in$ \{Cecum, Ascending Colon, Transverse Colon, Descending Colon, Sigmoid Colon, Rectum\}. The individual confusion matrices from colonoscopy videos in the test set were then averaged to build the final confusion matrix of colon segment classification. Additionally, the F1 score, sensitivity, precision, specificity, and accuracy of the individual colon segment classifications were calculated. 

The second method for evaluating the location index is to compare the trajectory of the location index from camera motion estimation and that from ScopeGuide length. As the Scope-Guide length can accurately indicate the distance that the camera travels when no loops are generated, one should expect a similar pattern in these two trajectories.

\section{Results and Discussion}
\subsection{Image classification performance}

\begin{table*}[]
\small
\centering
\begin{tabular}{@{}ccccccc@{}}
\toprule
Model                                & AUCPR         & AUC           & F1            & Sensitivity   & Specificity   & Precision     \\ \midrule
Non-informative image classification & 0.889 (0.058) & 0.959 (0.014) & 0.824 (0.095) & 0.801 (0.084) & 0.956 (0.026) & 0.732 (0.064) \\
biopsy forceps detection  & 0.942 (0.048) & 0.988 (0.010) & 0.848 (0.077) & 0.917 (0.076) & 0.983 (0.015) & 0.801 (0.117) \\ \bottomrule
\end{tabular}
\caption{Image classification performance. Average value and standard deviation across colonoscopy videos are provided.}
\label{tab:image_classification}
\end{table*}

Table \ref{tab:image_classification} shows the performance of non-informative image classification and biopsy forceps detection. From the table, both of the image classification models achieved high AUC and AUCPR. The trained image classification models will be used for the videos in the localization dataset.

\subsection{Visualization of overall framework}
\begin{figure*}[htp] 
\centering
\includegraphics[width=6.8in]{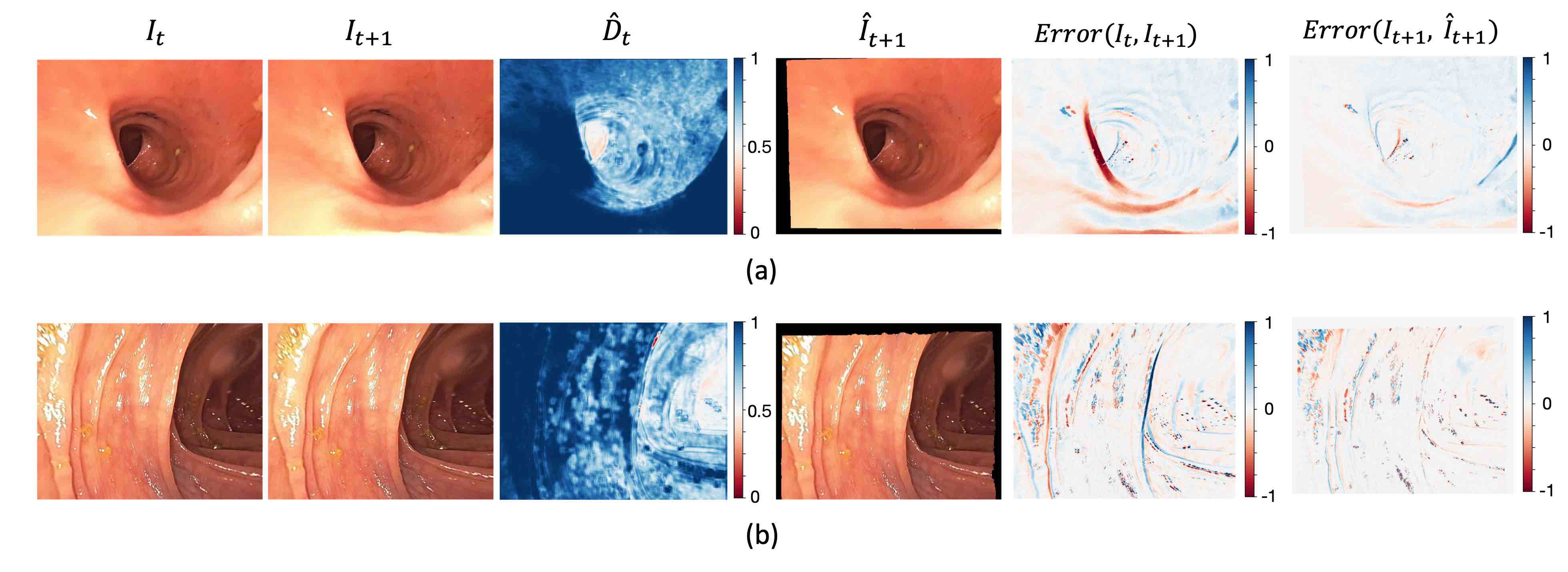}
\caption{Examples of estimated disparity maps, synthesized frames, and corresponding errors. $I_t$ and $I_{t+1}$ are the pair of input fed into the motion estimation network; $\hat D_t$ is the estimated disparity map for $I_t$; and $\hat I_{t+1}$ is the synthesised frame given $I_t$, $\hat D_t$, and $\hat s_{t\to t+1}$. $Error(I_t, I_{t+1})$ shows the photometric difference map between $I_t$ and $I_{t+1}$.  $Error(\hat I_{t+1}, I_{t+1})$ shows the difference map between the synthesised frame and the real frame.}
\label{fig:estimation_example}
\end{figure*}

Figure \ref{fig:estimation_example} gives examples of input images, estimated disparity maps, synthesized frames, and error maps from the trained model to illustrate the motion estimation algorithm. Figure \ref{fig:estimation_example} (a) presents an example of input frames with thin textural feature. Figure \ref{fig:estimation_example} (b) presents an example of input frames with many specular reflections. 

$I_t$ was fed into the disparity network to generate $\hat D_t$. Comparing $I_t$ and $\hat D_t$ in Figure \ref{fig:estimation_example}, one can observe the coordinates in the colon lumen were estimated with a lower disparity value, which is consistent with the fact that they are farther from the camera. In contrast, the coordinates in the colon wall, which are close to the camera, have a disparity value near 1. One can also observe that the specular regions impair the disparity map as those regions are very bright; as such, there is no information to infer the disparity.

\begin{figure*}[htp] 
\centering
\includegraphics[width=7in]{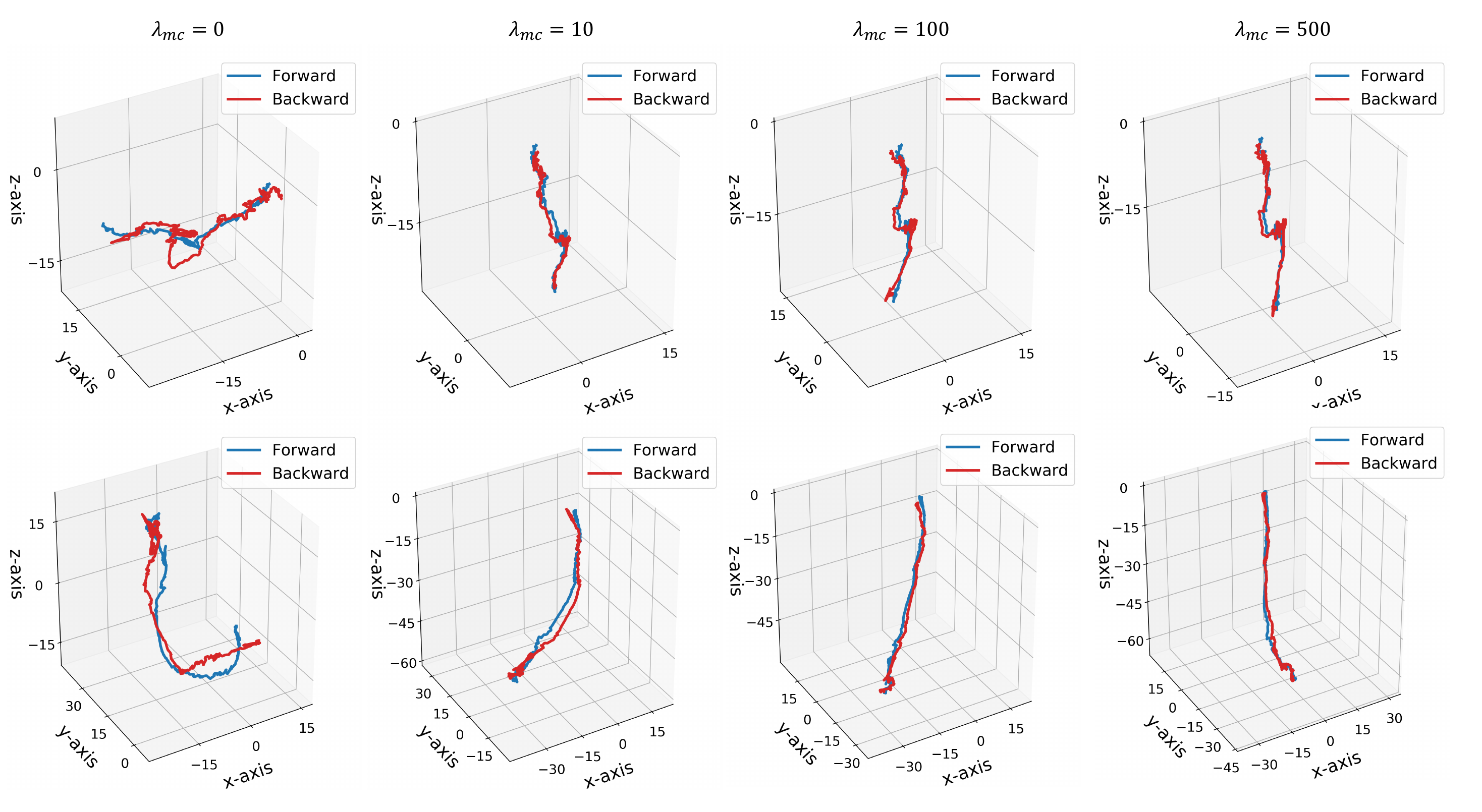}
\caption{The comparison of camera trajectories computed using $\hat s_{t\to t+1}$ (``forward'') and using $\hat s_{t+1\to t}$ (``backward''). Each column presents trajectories from a model trained with a different $\lambda$ value.}
\label{fig:consistency}
\end{figure*}

\begin{table*}[]
\small
\begin{adjustbox}{width=500pt, center}
\begin{tabular}{@{}c|ccc|ccc|ccc@{}}
\toprule
\multirow{2}{*}{Model} & \multicolumn{3}{c|}{Video \#1}                                                                                                                                         & \multicolumn{3}{c|}{Video \#2}                                                                                                                                        & \multicolumn{3}{c}{Video \#3}                                                                                                                                         \\
                       & \begin{tabular}[c]{@{}c@{}}CPE \\ ($\times1{\rm e}^{-2}$)\end{tabular} & \begin{tabular}[c]{@{}c@{}}RPE Trans.\\ ($\times1{\rm e}^{-2}$)\end{tabular} & \begin{tabular}[c]{@{}c@{}}RPE Rot.\\ ($\times1{\rm e}^{-2}\ ^{\circ} $)\end{tabular} & \begin{tabular}[c]{@{}c@{}}CPE\\ ($\times1{\rm e}^{-2}$)\end{tabular} & \begin{tabular}[c]{@{}c@{}}RPE Trans.\\ ($\times1{\rm e}^{-2}$)\end{tabular} & \begin{tabular}[c]{@{}c@{}}RPE Rot.\\ ($\times1{\rm e}^{-2}\ ^{\circ} $)\end{tabular} & \begin{tabular}[c]{@{}c@{}}CPE\\ (1$\times1{\rm e}^{-2})$\end{tabular} & \begin{tabular}[c]{@{}c@{}}RPE Trans.\\ ($\times1{\rm e}^{-2}$)\end{tabular} & \begin{tabular}[c]{@{}c@{}}RPE Rot.\\ ($\times1{\rm e}^{-2}\ ^{\circ} $)\end{tabular} \\ \midrule
$\lambda_{mc}=0$               & 6.09 (4.55)                                                            & 3.43 (6.75)                                                                  & 1.38 (2.85)                                                                 & 5.85 (3.77)                                                           & 5.80 (6.03)                                                                  & 1.16 (1.25)                                                                 & 5.49 (4.34)                                                           & 3.15 (3.80)                                                                  & 0.94 (1.73)                                                                \\
$\lambda_{mc}=10$               & 6.10 (5.05)                                                            & 1.41 (1.76)                                                                  & 0.26 (0.27)                                                                 & \textbf{5.66 (3.99)}                                                  & 2.29 (1.89)                                                                  & 0.26 (0.21)                                                                 & 5.30 (4.56)                                                           & 1.03 (1.18)                                                                  & 0.24 (0.21)                                                                \\
$\lambda_{mc}=100$               & \textbf{6.06 (5.02)}                                                   & 1.09 (1.58)                                                                  & 0.13 (0.11)                                                                 & \textbf{5.66 (4.04)}                                                  & 1.61 (1.46)                                                                  & 0.10 (0.07)                                                                 & \textbf{5.28 (4.60)}                                                  & 0.77 (1.03)                                                                  & 0.10 (0.07)                                                                \\
$\lambda_{mc}=500$              & 6.27 (5.05)                                                            & 1.20 (1.37)                                                                  & 0.06 (0.04)                                                                 & 5.90 (4.07)                                                           & 1.23 (1.08)                                                                  & 0.05 (0.03)                                                                 & 5.48 (4.60)                                                           & 0.63 (0.97)                                                                  & 0.05 (0.03)                                                                                                                          \\ \hline
\end{tabular}
\end{adjustbox}
\caption{Evaluation of the movement consistency term on the test set from the localization dataset Set 1. The standard deviation (std) was calculated over pairs of consecutive frames. The results are presented in the format of mean (std).}
\label{tab:consistency}
\end{table*}

With $I_t$, $\hat D_t$ and $\hat s_{t\to t+1}$, the frame at $t+1$ can be synthesized as $\hat I_{t+1}$. With an accurate estimation, $\hat I_{t+1}$ should be closer to the real frame $I_{t+1}$ than $I_t$. In Figure \ref{fig:estimation_example}, the difference map between the original frames $I_t$ and $I_{t+1}$, $Error(I_t, I_{t+1})$, is compared with the difference map between $\hat I_{t+1}$ and $I_{t+1}$: $Error(\hat I_{t+1}, I_{t+1})$. We can observe that the absolute photometric difference around the edges is quite high in $Error(I_t, I_{t+1})$ because of the camera's motion; those values are largely reduced in $Error(\hat I_{t+1}, I_{t+1})$. In addition, one can observe that the absolute photometric difference in specular regions is
high in both $Error(I_t, I_{t+1})$, and $Error(\hat I_{t+1}, I_{t+1})$, which indicates the necessity of using specular masks for loss calculation.

\subsection{Evaluation of the proposed movement consistency term}

The training and validation sets of the localization dataset Set 1 were used to find the optimal value for $\lambda_{mc}$. From our experimental results, $\lambda_{mc}=100$ leads to the best motion estimation performance. To further explore the consistency issue, models trained with $\lambda \in \{0, 10, 100, 500\}$ were tested on the test set. $CPE$ was calculated to evaluate the motion estimation performance. With trained models, a ``forward'' trajectory was derived using $\hat T_{t\to t+1}$ (i.e., the transformation matrix for $\hat s_{t\to t+1}$) and a ``backward'' trajectory was derived using the inverse of $\hat T_{t+1\to t}$. RPE Trans. and RPE Rot. were calculated to evaluate the consistency between $\hat T_{t\to t+1}$ and $\hat T_{t+1 \to t}$. If the equation (\ref{eq:consistency}) holds, we should have small RPE Trans. and RPE Rot. values. 

The evaluation results are shown in Table \ref{tab:consistency}. The visualizations of the ``forward'' trajectory and ``backward'' trajectory for two videos are shown in Figure \ref{fig:consistency}, where the ``forward'' trajectory and ``backward'' trajectory were aligned by a similarity transformation for a better comparison. Each column in Figure \ref{fig:consistency} presents camera trajectories derived from a model trained with different $\lambda_{mc}$. As mentioned in Section \ref{sec:related_work}, the pose estimation from the network is of arbitrary scale. The magnitude of $loss_{mc}$ can affect the scale of the estimated pose. As such, for each $\lambda_{mc}$, the estimated camera pose was scaled by a factor that constrains the length of estimated ``forward'' trajectory for the first video in the test set to 10. The scaling factor was used for a fair comparison among models trained with different $\lambda_{mc}$ values.

\begin{table}[]
\small
\centering
\begin{tabular}{@{}ccc@{}}
\toprule
                          & Train           & Test            \\ \midrule
Proposed                  & 0.0490 (0.0090) & 0.0567 (0.0032) \\
with fixed specular mask  & 0.0510 (0.0088) & 0.0580 (0.0033) \\
w/o specular mask         & 0.0496 (0.0091) & 0.0593 (0.0037) \\
w/o optical flow as input & 0.0485 (0.0073) & 0.0591 (0.0039) \\
w/o consistency term      & 0.0488 (0.0079) & 0.0581 (0.0025) \\ \bottomrule
\end{tabular}
\caption{Corrected photometric error (CPE) on the training and test set of localization dataset Set 1.}
\label{tab:cpe_evaluation}
\end{table}

\begin{table*}[!b]
\begin{adjustbox}{width=500pt, center}
\begin{tabular}{@{}c|ccc|ccc|ccc@{}}
\toprule
\multirow{2}{*}{Model}    & \multicolumn{3}{c|}{Colon-IV Trajectory-2}                                                                                                                            & \multicolumn{3}{c|}{Small Intestine-IV Trajectory-1}                                                                                                                  & \multicolumn{3}{c}{Stomach-II Trajectory-4}                                                                                                                           \\
                          & \begin{tabular}[c]{@{}c@{}}ATE\\ ($\times1{\rm e}^{-2}$)\end{tabular} & \begin{tabular}[c]{@{}c@{}}RPE Trans.\\ ($\times1{\rm e}^{-2}$)\end{tabular} & \begin{tabular}[c]{@{}c@{}}RPE Rot.\\ ($\times1{\rm e}^{-2}\ ^{\circ} $)\end{tabular} & \begin{tabular}[c]{@{}c@{}}ATE\\ ($\times1{\rm e}^{-2}$)\end{tabular} & \begin{tabular}[c]{@{}c@{}}RPE Trans.\\ ($\times1{\rm e}^{-2}$)\end{tabular} & \begin{tabular}[c]{@{}c@{}}RPE Rot.\\ ($\times1{\rm e}^{-2}\ ^{\circ} $)\end{tabular} & \begin{tabular}[c]{@{}c@{}}ATE\\ ($\times1{\rm e}^{-2}$)\end{tabular} & \begin{tabular}[c]{@{}c@{}}RPE Trans.\\ ($\times1{\rm e}^{-2}$)\end{tabular} & \begin{tabular}[c]{@{}c@{}}RPE Rot.\\ ($\times1{\rm e}^{-2}\ ^{\circ} $)\end{tabular} \\ \midrule
Proposed                                    & \textbf{1.98 (0.61)}                                                  & \textbf{0.16 (0.16)}                                                         & \textbf{0.85 (0.83)}                                                        & \textbf{2.82 (1.14)}                                                  & 0.22 (0.15)                                                                  & \textbf{0.83 (0.74)}                                                        & \textbf{2.63 (1.36)}                                                  & 0.33 (0.23)                                                                  & \textbf{0.91 (0.69)}                                                       \\
with fixed specular mask                    & 2.22 (0.77)                                                           & \textbf{0.16 (0.16)}                                                         & 0.98 (0.84)                                                                 & 3.82 (1.66)                                                           & 0.21 (0.15)                                                                  & 0.96 (0.74)                                                                 & 3.62 (1.69)                                                           & 0.32 (0.21)                                                                  & 0.96 (0.69)                                                                \\
w/o specular mask                           & 2.99 (1.31)                                                           & 0.17 (0.16)                                                                  & 1.03 (0.85)                                                                 & 3.52 (1.66)                                                           & 0.22 (0.15)                                                                  & 1.09 (0.73)                                                                 & 3.82 (1.68)                                                           & 0.32 (0.24)                                                                  & 1.06 (0.68)                                                                \\
w/o optical flow as input                   & 2.25 (0.87)                                                           & 0.18 (0.17)                                                                  & 1.56 (1.04)                                                                 & 5.27 (1.88)                                                           & \textbf{0.20 (0.12)}                                                         & 1.70 (1.11)                                                                 & 4.05 (1.78)                                                           & 0.33 (0.23)                                                                  & 1.33 (1.44)                                                                \\
w/o consistency term                        & 2.98 (1.27)                                                           & 0.20 (0.21)                                                                  & 2.93 (3.89)                                                                 & 5.77 (2.17)                                                           & 0.22 (0.18)                                                                  & 3.76 (4.32)                                                                 & 4.11 (1.69)                                                           & 0.35 (0.31)                                                                  & 4.58 (6.31)                                                                \\ \hline
SfMLearner \citep{zhou2017unsupervised}                                        & 2.98 (1.15)                                                           & 0.24 (0.20)                                                                  & 1.40 (0.97)                                                                 & 5.09 (1.97)                                                           & 0.21 (0.14)                                                                  & 1.05 (0.81)                                                                 & 4.28 (1.54)                                                           & 0.37 (0.31)                                                                  & 2.28 (2.28)                                                                \\
GeoNet     \citep{yin2018geonet}                                   & 2.75 (1.08)                                                           & 0.18 (0.16)                                                                  & 1.32 (0.89)                                                                 & 4.31 (1.77)                                                           & 0.21 (0.12)                                                                  & 1.38 (0.83)                                                                 & 4.26 (1.59)                                                           & 0.34 (0.22)                                                                  & 1.34 (0.80)                                                                \\
SC-SfMLearner       \citep{bian2019unsupervised}                               & 2.96 (1.08)                                                           & 0.21 (0.19)                                                                  & 1.08 (0.79)                                                                 & 4.23 (2.11)                                                           & 0.21 (0.16)                                                                  & 0.94 (0.70)                                                                 & 4.25 (1.58)                                                           & 0.34 (0.29)                                                                  & 1.04 (0.74)                                                                \\
optical flow-based \citep{liu2013optical}                         & 3.15 (1.38)                                                           & 0.19 (0.19)                                                                  & 4.97 (4.98)                                                                 & 4.89 (2.43)                                                           & \textbf{0.20 (0.15)}                                                         & 5.34 (4.27)                                                                 & 4.18 (1.71)                                                           & \textbf{0.30 (0.25)}                                                         & 6.09 (5.34)                                                                \\ \hline
\end{tabular}
\end{adjustbox}
\caption{Performance comparison of motion estimation algorithms on high-resolution videos from EndoSLAM dataset. The standard deviation (std) was calculated over pairs of the consecutive frames. The results are presented in the format of mean (std). ``Proposed'': the proposed method; ``with fixed specular mask'': the proposed method with a corrected photometric loss using a threshold-based specular mask; ``w/o specular mask'':  the proposed method without applying the specular mask to correct the photometric loss; ``w/o optical flow as input'': the proposed method without using optical flow as input; ``w/o consistency term'': the proposed method without adding the movement consistency term.}
\label{tab:high-resolution}
\end{table*}

\begin{table*}
\begin{adjustbox}{width=500pt, center}
\begin{tabular}{@{}c|ccc|ccc|ccc@{}}
\toprule
\multirow{2}{*}{Model}    & \multicolumn{3}{c|}{High-resolution videos}                                                                                                                            & \multicolumn{3}{c|}{Low-resolution videos}                                                                                                                  & \multicolumn{3}{c}{All videos}                                                                                                                           \\
                          & \begin{tabular}[c]{@{}c@{}}ATE\\ ($\times1{\rm e}^{-2}$)\end{tabular} & \begin{tabular}[c]{@{}c@{}}RPE Trans.\\ ($\times1{\rm e}^{-2}$)\end{tabular} & \begin{tabular}[c]{@{}c@{}}RPE Rot.\\ ($\times1{\rm e}^{-2}\ ^{\circ} $)\end{tabular} & \begin{tabular}[c]{@{}c@{}}ATE\\ ($\times1{\rm e}^{-2}$)\end{tabular} & \begin{tabular}[c]{@{}c@{}}RPE Trans.\\ ($\times1{\rm e}^{-2}$)\end{tabular} & \begin{tabular}[c]{@{}c@{}}RPE Rot.\\ ($\times1{\rm e}^{-2}\ ^{\circ} $)\end{tabular} & \begin{tabular}[c]{@{}c@{}}ATE\\ ($\times1{\rm e}^{-2}$)\end{tabular} & \begin{tabular}[c]{@{}c@{}}RPE Trans.\\ ($\times1{\rm e}^{-2}$)\end{tabular} & \begin{tabular}[c]{@{}c@{}}RPE Rot.\\ ($\times1{\rm e}^{-2}\ ^{\circ} $)\end{tabular} \\ \midrule

Proposed                                    & \textbf{2.48 (0.36)}                                                  & 0.24 (0.07)                                                                  & \textbf{0.86 (0.03)}                                                        & \textbf{2.36 (0.30)}                                                  & 0.34 (0.12)                                                                  & \textbf{1.10 (0.24)}                                                       & \textbf{2.42 (0.33)}                                                  & 0.29 (0.11)                                                                  & \textbf{0.98 (0.21)}                                                       \\
with fixed specular mask                    & 3.22 (0.71)                                                           & \textbf{0.23 (0.07)}                                                         & 0.97 (0.01)                                                                 & 2.44 (0.64)                                                           & 0.34 (0.12)                                                                  & 1.13 (0.22)                                                                 & 2.73 (0.84)                                                           & \textbf{0.28 (0.11)}                                                         & 1.05 (0.17)                                                                \\
w/o specular mask                           & 3.44 (0.34)                                                           & 0.24 (0.07)                                                                  & 1.06 (0.02)                                                                 & 3.14 (0.94)                                                           & 0.34 (0.12)                                                                  & 1.29 (0.25)                                                                 & 3.29 (0.72)                                                           & 0.29 (0.11)                                                                  & 1.18 (0.21)                                                                \\
w/o optical flow as input                   & 4.04 (1.43)                                                           & 0.24 (0.06)                                                                  & 1.53 (0.15)                                                                 & 3.03 (0.69)                                                           & 0.34 (0.12)                                                                  & 1.46 (0.66)                                                                 & 3.54 (1.23)                                                           & 0.29 (0.11)                                                                  & 1.50 (0.48)                                                                \\
w/o consistency term                        & 4.29 (1.15)                                                           & 0.26 (0.06)                                                                  & 3.76 (0.67)                                                                 & 3.25 (0.99)                                                           & 0.34 (0.11)                                                                  & 3.22 (1.77)                                                                 & 3.77 (1.19)                                                           & 0.30 (0.10)                                                                  & 3.49 (1.37)                                                                \\ \hline
SfMLearner     \citep{zhou2017unsupervised}                                    & 4.12 (0.87)                                                           & 0.27 (0.07)                                                                  & 1.58 (0.52)                                                                 & 3.98 (0.97)                                                           & 0.36 (0.12)                                                                  & 2.99 (0.46)                                                                 & 4.05 (0.93)                                                           & 0.32 (0.11)                                                                  & 2.28 (0.86)                                                                \\
GeoNet \citep{yin2018geonet}                                      & 3.77 (0.72)                                                           & 0.24 (0.07)                               
& 1.35 (0.02)                                                                 & 3.56 (0.95)                                                           & 0.34 (0.11)                                                                  & 1.25 (0.28)                                                                 & 3.67 (0.85)                                                           & 0.29 (0.11)                                                                  & 1.30 (0.20)                                                                \\
SC-SfMLearner \citep{bian2019unsupervised}                                    & 3.81 (0.60)                                                           & 0.25 (0.06)                                                                  & 1.02 (0.06)                                                                 & 3.78 (0.92)                                                           & 0.35 (0.10)                                                                  & 1.12 (0.16)                                                                & 3.80 (0.78)                                                           & 0.30 (0.10)                                                                  & 1.05 (0.12)                                                                \\
optical flow-based  \citep{liu2013optical}                      & 4.07 (0.72)                                                           & \textbf{0.23 (0.05)}                                                         & 5.47 (0.47)                                                                 & 4.05 (0.75)                                                           & \textbf{0.33 (0.08)}                                                         & 6.11 (0.95)                                                                 & 4.06 (0.73)                                                           & \textbf{0.28 (0.08)}                                                         & 5.79 (0.81)                                                                \\ \hline
\end{tabular}
\end{adjustbox}
\caption{Performance comparison of motion estimation algorithms on the EndoSLAM dataset. The standard deviation (std) was calculated over videos. The results are presented in the format of mean (std).}
\label{tab:overall_endoslam}
\end{table*}

We can observe that when $\lambda_{mc} = 0$, the ``forward'' trajectory differs from the ``backward'' trajectory, and RPE Trans. and RPE Rot. are the highest. This indicates that equation (\ref{eq:consistency}) doesn't hold. With an increasing $\lambda_{mc}$, the ``forward'' trajectory becomes closer to the ``backward'' trajectory, and RPE Trans. and RPE Rot. become lower. From Table \ref{tab:consistency}, $\lambda_{mc}=100$ leads to a lower CPE for all three videos when compared with $\lambda_{mc}=0$. Our experimental results show that the proposed movement consistency term can improve motion estimation performance by encouraging the consistency between $\hat T_{t\to t+1}$ and $\hat T_{t+1 \to t}$. The CPE decreases when $\lambda_{mc} = 500$. This is because $\lambda_{cm} loss_{cm}$ is too large and overwhelms $loss_{cp}$ in equation (\ref{eq:final_loss}).

\begin{figure*}[] 
\centering
\includegraphics[width=7in]{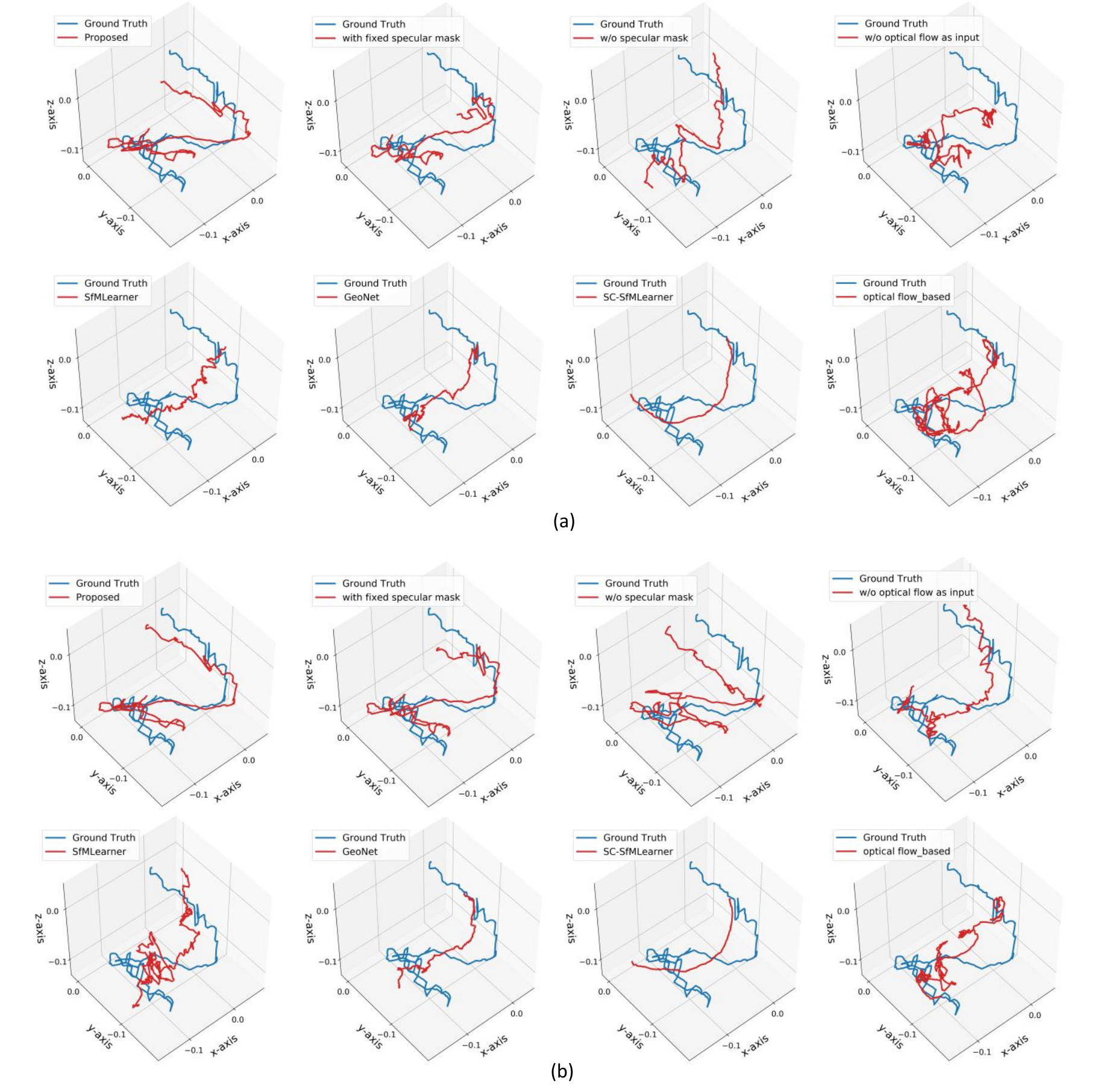}
\caption{Comparison of the estimated trajectories and ground truth trajectories. (a) Trajectories for the high-resolution video on Small Intestine-IV Trajectory-1. (b) Trajectories for the low-resolution video on Small Intestine-IV Trajectory-1.}
\label{fig:endoslam}
\end{figure*}

\subsection{Evaluation of the proposed camera pose estimation method}
For the proposed camera pose estimation method, three major modifications were made, including correcting photometric loss with an estimated specular mask, adding optical flow as input, and adding movement consistency terms. In Table \ref{tab:cpe_evaluation}, the proposed model and the proposed method without one of these modifications were evaluated on the training and test set of the localization dataset Set 1. Models presented includes the proposed method (``Proposed''), the proposed method with a corrected photometric loss using a threshold-based specular mask (``with fixed specular mask''), the proposed method without applying the specular mask to correct the photometric loss (``w/o specular mask''), the proposed method without using optical flow as input (``w/o optical flow as input''), and the proposed method without adding movement consistency term (``w/o consistency term'').

Further, we evaluated the proposed camera pose estimation on the EndoSLAM dataset. The EndoSLAM dataset is an external dataset with ground truth for camera pose. Besides the aforementioned models, four existing techniques: ``SfMLearner''\citep{zhou2017unsupervised}, ``GeoNet''\citep{yin2018geonet}, ``SC-SfMLearner''\citep{bian2019unsupervised},  and ``optical flow-based'' \citep{liu2013optical} from recent literature with publicly available code were also evaluated on the EndoSLAM dataset for comparison. 

Figure \ref{fig:endoslam} compares the estimated trajectory and ground truth trajectory on Small Intestine-IV Trajectory-1. We can observe that the estimated trajectories from the proposed method are the most accurate with loops of similar shape as compared to the ground truth trajectories. In addition, the trajectory estimated on the high-resolution video is very close to the trajectory on the low-resolution video, which indicates good generalization of the proposed method. The ``with fixed specular mask'' model also achieved good performance. However, compared with its estimated trajectory on low-resolution video, the accuracy of the estimated trajectory on high-resolution video decreases.

Quantitative measurements were calculated and are shown in Tables \ref{tab:high-resolution} and \ref{tab:overall_endoslam}. Table \ref{tab:high-resolution} presents the performance of models on individual high-resolution videos. Table \ref{tab:overall_endoslam} summarizes the average performance on high-resolution videos, low-resolution videos, and the entire dataset. The proposed method achieved the lowest ATE and RPE Rot. The RPE Trans. values for all methods are quite close. While the ``optical flow-based'' model has the lowest RPE Trans., its ATE and RPE Rot. are very high. From our experimental results, the proposed method achieves the best performance on the EndoSLAM dataset.

\begin{figure*}[!ht] 
\centering
\includegraphics[width=7.5in]{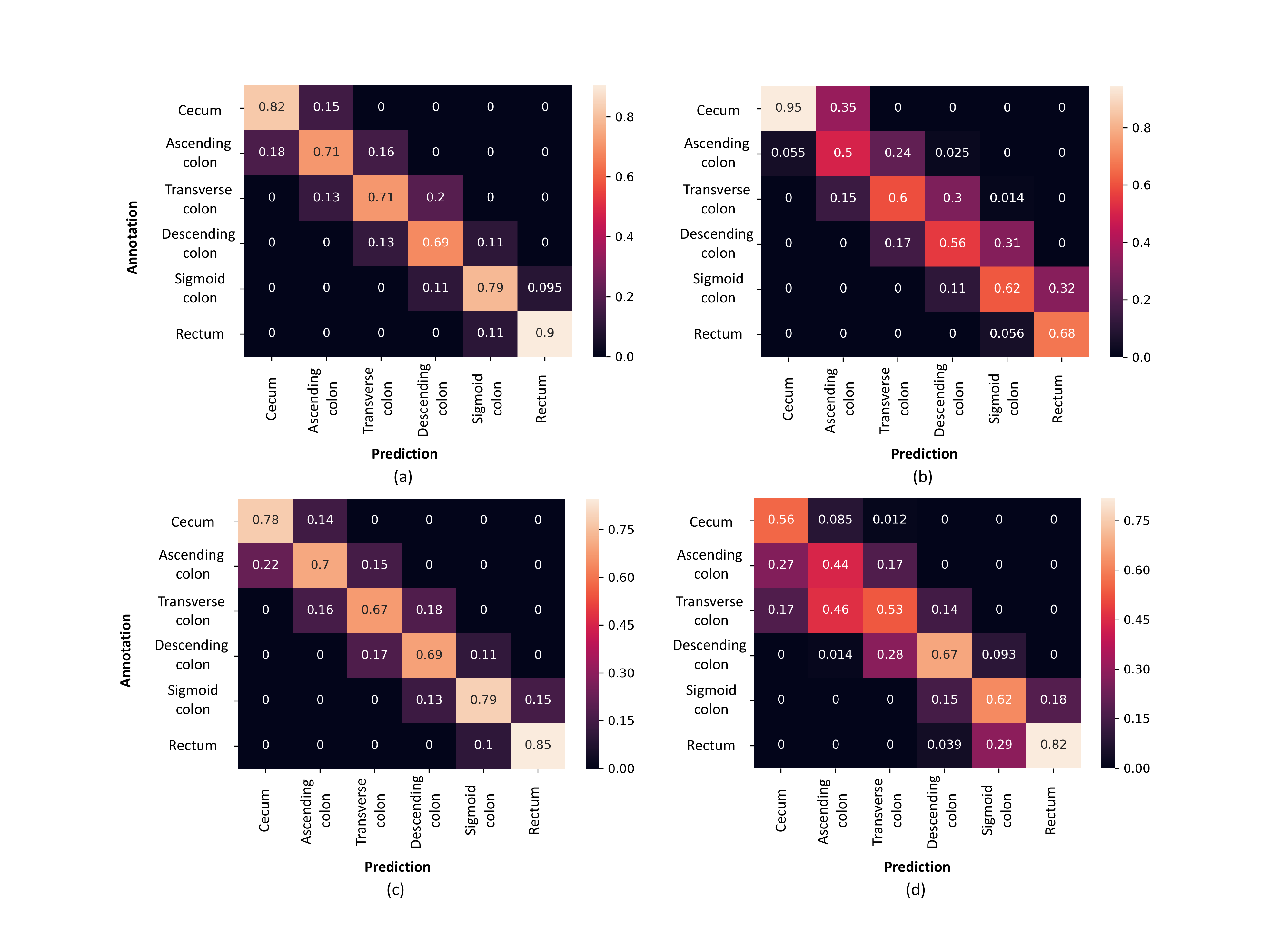}
\caption{(a)-(b): Confusion matrix between the classified colon segment and annotated colon segment on the entire test set. (a) is the result of using the location index, and (b) is the result of using the time index. (c)-(d): Confusion matrix between the classified colon segment and annotated colon segment from leave-one-out cross-validation on the test set. (c) is the result of using the location index, and (d) is the result of using length from ScopeGuide videos. The calculation of the confusion matrix is described in Section \ref{sec:evaluation_strategy}. }
\label{fig:cm_1}
\end{figure*}

\begin{table*}[]
\small
\centering
\begin{tabular}{@{}ccccccc@{}}
\toprule
                                              Method   & Cecum & \begin{tabular}[c]{@{}c@{}}Ascending \\ Colon\end{tabular}  & \begin{tabular}[c]{@{}c@{}}Transverse\\ Colon\end{tabular} & \begin{tabular}[c]{@{}c@{}}Descending\\ Colon\end{tabular} & \begin{tabular}[c]{@{}c@{}}Sigmoid \\ Colon\end{tabular} & Rectum \\ \midrule
Motion-based location index (Set 2)              & 0.061  & 0.146           & 0.224            & 0.223            & 0.258         & 0.088  \\
Time index (Set 2)                               & 0.068  & 0.217           & 0.210             & 0.177            & 0.172         & 0.156  \\
Motion-based location index  (Set 3)             & 0.067  & 0.142           & 0.245            & 0.204            & 0.244         & 0.096  \\
ScopeGuide Length-based location index   (Set 3) & 0.236  & 0.129           & 0.162            & 0.204            & 0.167         & 0.102  \\ \bottomrule
\end{tabular}
\caption{Colon templates estimated using different strategies. In the first two rows, the colon templates were built on Set 2 of the localization dataset. In the last two rows, the colon templates were the average of those from cross-validation on Set 3 of the localization dataset.}
\label{tab:template}
\end{table*}

\begin{table}[!t]
\small
\begin{adjustbox}{width=250pt, center}
\begin{tabular}{@{}cccc@{}}
\toprule
                                                                                            & Accuracy      & \begin{tabular}[c]{@{}c@{}}Average \\ segment error\end{tabular} & \begin{tabular}[c]{@{}c@{}}Max.\\ segment error\end{tabular} \\ \midrule
\begin{tabular}[c]{@{}c@{}}Motion-based \\ location index (Set 2)\end{tabular}              & 0.754 (0.111) & 0.246 (0.111)                                                     & 1.0 (0.0)                                                       \\ \midrule
\begin{tabular}[c]{@{}c@{}}Time index  (Set 2)\end{tabular}                               & 0.608 (0.140) & 0.399 (0.151)                                                     & 1.2 (0.4)                                                       \\ \midrule
\begin{tabular}[c]{@{}c@{}}Motion-based \\ location index (Set 3)\end{tabular}             & 0.718 (0.097) & 0.282 (0.097)                                                     & 1.0 (0.0)                                                       \\ \midrule
\begin{tabular}[c]{@{}c@{}}ScopeGuide \\Length-based \\ location index (Set 3)\end{tabular} & 0.587 (0.134) & 0.472 (0.172)                                                     & 1.9 (0.3)                                                       \\ \midrule
\begin{tabular}[c]{@{}c@{}} Motion-based \\location index (Set 2) \\ with all frames\end{tabular} & 0.576 (0.117) & 0.437 (0.131)                                                     & 1.2 (0.4)                                                       \\ \midrule
\begin{tabular}[c]{@{}c@{}}Motion-based \\location index (Set 3) \\ with all frames \end{tabular} & 0.579 (0.175) & 0.436 (0.189)                                                     & 1.2 (0.4)                                                       \\

\bottomrule
\end{tabular}
\end{adjustbox}
\caption{Classification accuracy and segment errors. In the first two rows and the last two rows, the colon templates were built on Set 2 of the localization dataset. In the middle two rows, the colon templates were the average of those from cross-validation on Set 3 of the localization dataset. In the last two rows, the camera trajectories were estimated using all frames from colonoscopy videos, meaning non-informative frames and frames with biopsy forceps were not removed.}
\label{tab:evaluation}

\end{table}

\subsection{Anatomical colon segment classification performance}

When comparing the performance of the colon segment classification using motion-based location index and using the time index, Set 2 of the localization dataset was used for template building and Set 3 of the localization dataset was used for testing. 

The colon template built from manual time annotations and the location index from our localization system are shown in the first row of Table \ref{tab:template}, wherein each entry corresponds to the relative length of each colon segment (from cecum to rectum). From the estimated colon template, the cecum and rectum are shorter, occupying less than 10\% of the total colon length, while the other four colon segments are longer, occupying 15\% to 20\% of the total colon length. The template is consistent with our physiological knowledge about the colon. We then performed colon segment classification on the test set.

The average accuracy and per-segment errors of the classification on the test set are given in Table \ref{tab:evaluation}. The confusion matrix of the classification is shown in Figure \ref{fig:cm_1} (a). The classification performance on individual colon segments is shown in Table \ref{tab:table_confusion_mat_1}. From the results, the anatomical colon segment classification is most accurate in the cecum and rectum and less accurate in the middle colon segments. This is an intrinsic property of the proposed classification method because the classification of the middle colon segments suffer from accumulated error. Also, one can observe that the frame will only be misclassified to the colon segments adjacent to the true segment.

For comparison, a colon template was built, and colon segment classification was performed using the time index. The colon template is shown in the second row of Table \ref{tab:template}. Using the time index, the relative lengths of the descending colon and sigmoid colon are similar to that of the rectum, which is not true. The average accuracy and per-segment errors of the classification are presented in Table \ref{tab:evaluation}. Both errors and their standard deviations are higher. The corresponding confusion matrix is shown in Figure \ref{fig:cm_1} (b), and individual colon segment classification performance is shown in Table \ref{tab:table_confusion_mat_2}. From these results, one can conclude that using the proposed location index is more accurate than using the time index.

We also compared the performance of anatomical colon segment classification using the camera motion-based location index with that using the ScopeGuide length-based location index. As only paired ScopeGuide videos were available in Set 3, leave-one-out cross-validation was used to evaluate the performance. For each fold, only one video was used for testing, and all others were used to build the colon template. The evaluation measurements were then averaged over all folds.

Using the motion-based location index, the average template from cross-validation on Set 1 is close to the template from Set 2, with around $\pm2$\% difference. The average accuracy and segment errors of the classification from cross-validation are shown in Table \ref{tab:evaluation}. Figure \ref{fig:cm_1} (c) and Table \ref{tab:table_confusion_mat_3} further show the cross-validation performance of using the motion-based location index. The overall classification performance is slightly lower than the performance in Figure \ref{fig:cm_1} (a) and Table \ref{tab:table_confusion_mat_1}.This may be due to the colon template being built from fewer cases. 

For colon segment classification using ScopeGuide length, the average accuracy and segment errors are presented in Table \ref{tab:evaluation}. Figure \ref{fig:cm_1} (d) and Table \ref{tab:table_confusion_mat_4} further show the performance from cross-validation. From these results, using the ScopeGuide length-based location index yields poor classification performance, much lower than that from the camera motion-based location index. This may be because the loops generated inside the body lead to sharp increases and decreases in length. This issue is discussed further in the next section. From the template built using the ScopeGuide length-based location index, the cecum segment is estimated to be more than 20\% of total colon length. The over-estimation of cecum length may result from the fact that loops are more easily generated when the amount of inserted scope is greater. The trajectory of the location index derived from ScopeGuide length is discussed in the next section.

\begin{table*}[hb]
\small
\centering
\begin{tabular}{@{}ccccccc@{}}
\toprule
            & Cecum         & Ascending Colon & Transverse Colon & Descending Colon & Sigmoid       & Rectum        \\ \midrule
F1          & 0.953 (0.035) & 0.901 (0.067)   & 0.874 (0.075)    & 0.878 (0.076)    & 0.927 (0.043) & 0.975 (0.033) \\
Sensitivity & 0.864 (0.239) & 0.696 (0.216)   & 0.693 (0.216)    & 0.750 (0.229)    & 0.786 (0.227) & 0.888 (0.257) \\
Specificity & 0.975 (0.021) & 0.949 (0.048)   & 0.928 (0.073)    & 0.915 (0.065)    & 0.957 (0.056) & 0.988 (0.016) \\
Precision   & 0.822 (0.171) & 0.714 (0.230)   & 0.713 (0.282)    & 0.686 (0.187)    & 0.785 (0.288) & 0.905 (0.139) \\
Accuaracy   & 0.953 (0.035) & 0.901 (0.067)   & 0.874 (0.075)    & 0.878 (0.076)    & 0.927 (0.043) & 0.975 (0.033) \\ \bottomrule
\end{tabular} 
\caption{Anatomical colon segment classification performance on Set 3 of the localization dataset using location index derived from camera motion estimation.}
\label{tab:table_confusion_mat_1}
\end{table*}

\begin{table*}[hb]
\small
\centering
\begin{tabular}{@{}ccccccc@{}}
\toprule
            & Cecum         & Ascending Colon & Transverse Colon & Descending Colon & Sigmoid       & Rectum        \\ \midrule
F1          & 0.919 (0.059) & 0.833 (0.114)   & 0.828 (0.077)    & 0.833 (0.063)    & 0.863 (0.038) & 0.940 (0.027) \\
Sensitivity & 0.566 (0.237) & 0.753 (0.324)   & 0.566 (0.220)    & 0.524 (0.211)    & 0.606 (0.141) & 0.962 (0.114) \\
Specificity & 0.996 (0.012) & 0.871 (0.074)   & 0.900 (0.073)    & 0.905 (0.043)    & 0.925 (0.055) & 0.945 (0.031) \\
Precision   & 0.945 (0.164) & 0.500 (0.266)   & 0.598 (0.303)    & 0.558 (0.217)    & 0.616 (0.294) & 0.675 (0.189) \\
Accuaracy   & 0.919 (0.059) & 0.833 (0.114)   & 0.828 (0.077)    & 0.833 (0.063)    & 0.863 (0.038) & 0.940 (0.027) \\ \bottomrule
\end{tabular}
\caption{Anatomical colon segment classification performance on Set 3 of the localization dataset using time index.}
\label{tab:table_confusion_mat_2}
\end{table*}

\begin{table*}[]
\small
\centering
\begin{tabular}{@{}ccccccc@{}}
\toprule
            & Cecum         & Ascending Colon & Transverse Colon & Descending Colon & Sigmoid       & Rectum        \\ \midrule
F1          & 0.940 (0.044) & 0.886 (0.069)   & 0.864 (0.074)    & 0.870 (0.069)    & 0.915 (0.046) & 0.963 (0.040) \\
Sensitivity & 0.873 (0.239) & 0.636 (0.226)   & 0.709 (0.219)    & 0.677 (0.225)    & 0.716 (0.222) & 0.890 (0.270) \\
Specificity & 0.959 (0.044) & 0.949 (0.046)   & 0.909 (0.075)    & 0.920 (0.060)    & 0.958 (0.056) & 0.976 (0.033) \\
Precision   & 0.776 (0.231) & 0.699 (0.227)   & 0.672 (0.276)    & 0.688 (0.197)    & 0.787 (0.292) & 0.847 (0.206) \\
Accuaracy   & 0.940 (0.044) & 0.886 (0.069)   & 0.864 (0.074)    & 0.870 (0.069)    & 0.915 (0.046) & 0.963 (0.040) \\ \bottomrule
\end{tabular}
\caption{Anatomical colon segment classification performance from leave-one-out cross-validation on Set 3 of the localization dataset using location index derived from camera motion estimation.}
\label{tab:table_confusion_mat_3}
\end{table*}

\begin{table*}[]
\small
\centering
\begin{tabular}{@{}ccccccc@{}}
\toprule
            & Cecum         & Ascending Colon & Transverse Colon & Descending Colon & Sigmoid       & Rectum        \\ \midrule
F1          & 0.855 (0.082) & 0.821 (0.065)   & 0.800 (0.103)    & 0.876 (0.066)    & 0.889 (0.063) & 0.934 (0.048) \\
Sensitivity & 0.893 (0.149) & 0.407 (0.338)   & 0.381 (0.226)    & 0.647 (0.250)    & 0.708 (0.284) & 0.590 (0.246) \\
Precision   & 0.558 (0.302) & 0.438 (0.324)   & 0.533 (0.251)    & 0.672 (0.249)    & 0.619 (0.238) & 0.818 (0.272) \\
Specificity & 0.856 (0.104) & 0.904 (0.071)   & 0.916 (0.051)    & 0.921 (0.061)    & 0.922 (0.042) & 0.984 (0.027) \\
Accuaracy   & 0.855 (0.082) & 0.821 (0.065)   & 0.800 (0.103)    & 0.876 (0.066)    & 0.889 (0.063) & 0.934 (0.048) \\ \bottomrule
\end{tabular}
\caption{Anatomical colon segment classification performance from leave-one-out cross-validation on Set 3 of the localization dataset using length from ScopeGuide videos.}
\label{tab:table_confusion_mat_4}
\end{table*}

\subsection{Camera trajectory comparison}
\begin{figure*}[!ht]  
\centering
\includegraphics[width=6.8in]{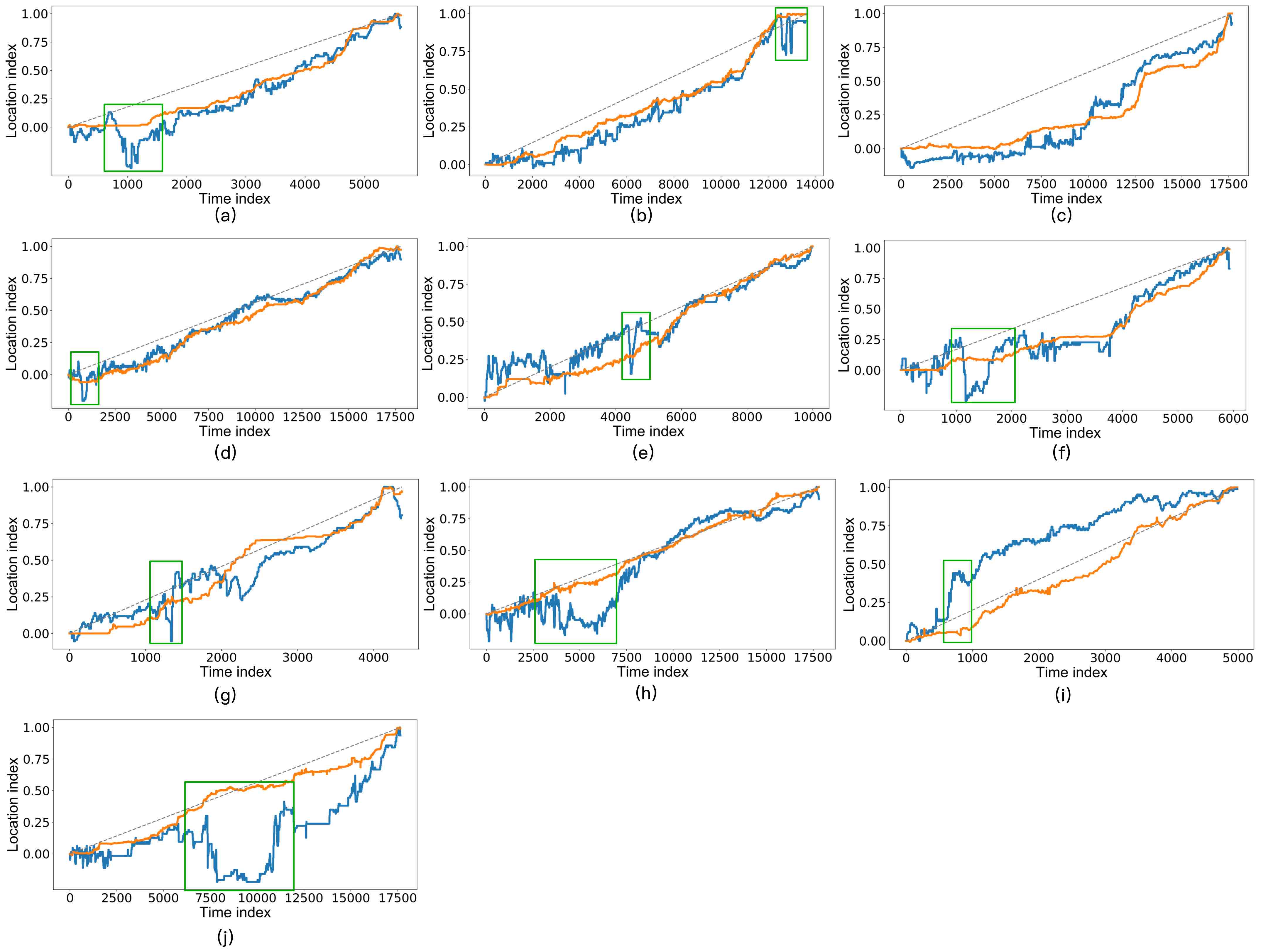}
\caption{Trajectories comparisons of location index calculated from different sources using Set 3 of the localization dataset. Orange line: location index derived from the camera motion estimation; Blue line: location index calculated using length in ScopeGuide videos; Dotted gray line: location index using the time index directly; Green box: regions with sharp increase/decrease in ScopeGuide length result from generated loops.}
\label{fig:tragectory}
\end{figure*}

Figure \ref{fig:tragectory} compares the location index calculated from camera motion estimation (orange line),  ScopeGuide length (blue line), and frame index (dotted gray line) of the 10 colonoscopy videos in Set 3 of the localization dataset.  From Figure \ref{fig:tragectory}, we can observe that the ScopeGuide length-based location index is rougher due to its low resolution. Also, sharp spikes or bumps can be observed. In those spikes (shown in green boxes), the location index may increase or decrease by 0.2 (usually equal to 15-25 cm) in a short time. A physician manually examined the paired colonoscopy videos and ScopeGuide videos, finding that those spikes or bumps occur because of loops generated when the colonoscopy performer advanced the colonoscope toward the cecum. In Figure \ref{fig:tragectory}(i), a sharp increase of the location index can be observed, which results from the loop generated in the insertion phase. Those spikes and bumps can significantly impair the proposed colon segment classification when the location index is derived from the ScopeGuide length, where the max segment error is 1.9 (0.3). Except for regions with shape spikes or bumps, the blue line and orange line have good consistency in the pattern of the location index sequence. It indicates that using a computer vision-based method can lead to a good sense of the camera's location. By comparing the dotted gray line with the other two lines, we can find sampling frames using the time index can lead to an unbalanced sampling of frames at different colon regions, emphasizing the importance of camera localization in colonoscopy video analysis.

\begin{figure}[!ht]
\centering
\includegraphics[width=3.2in]{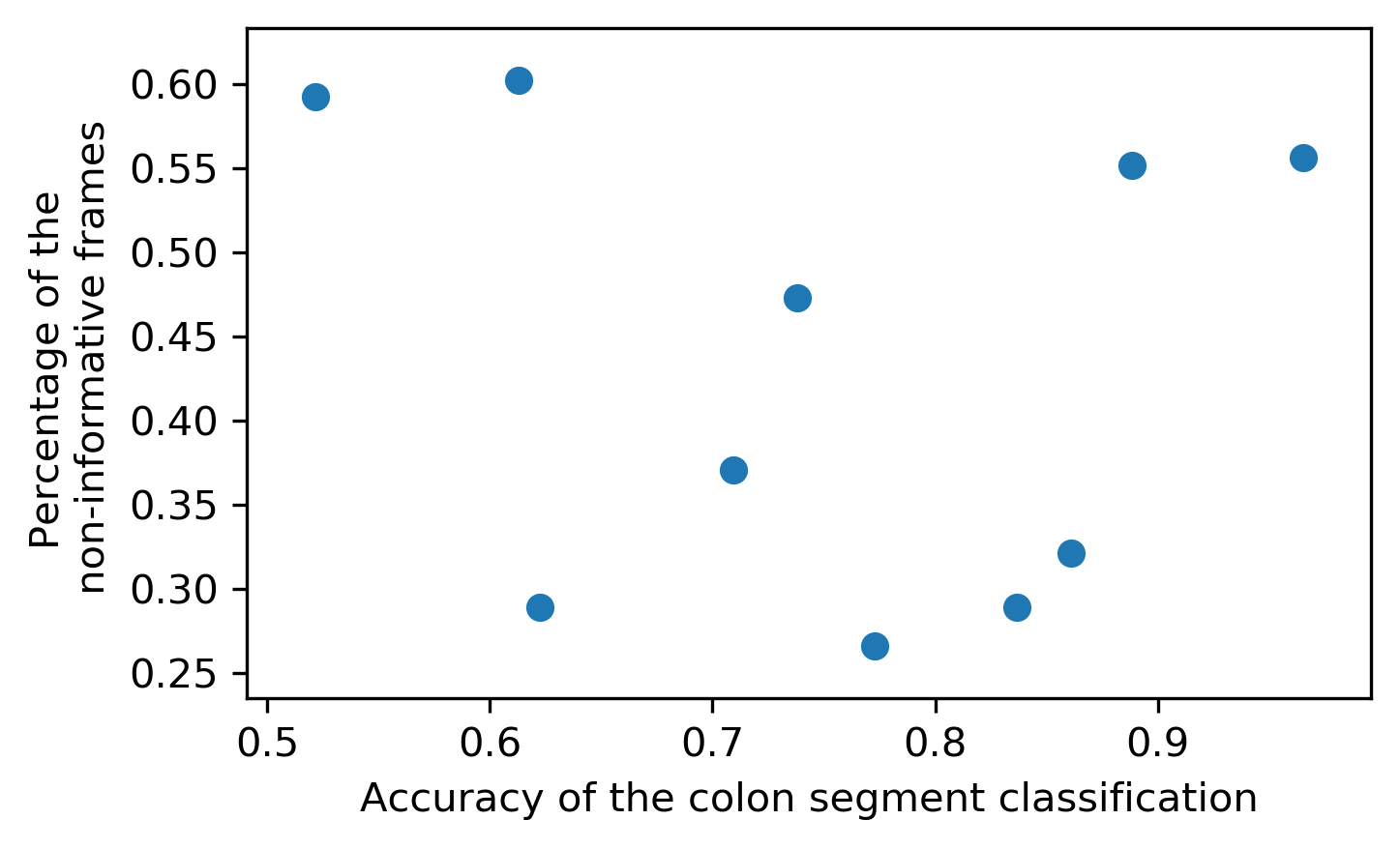}
\caption{A scatter plot of the percentage of non-informative frames versus colon segment classification accuracy.}
\label{fig:non_informative}
\end{figure}

\subsection{Interruption by non-informative frames}
In colonoscopy videos, the informative frame sequence can be interrupted by non-informative frames frequently. In the proposed method, non-informative frames were directly removed under the assumption that this would not affect the estimated location index significantly. In Figure \ref{fig:non_informative}, each point is a colono-scopy video in Set 3 of the localization dataset. The $y$-axis is the colon segment classification accuracy using the template built on Set 2, and the $x$-axis is the percentage of non-informative frames. From Figure \ref{fig:non_informative}, the accuracy of the colon segment classification is not related to the percentage of non-informative frames. 

To examine the necessity of removing non-informative frames and frames with biopsy forceps, we performed camera trajectory estimation and anatomical colon segment classification without removing non-informative and biopsy frames. The performance of the anatomical colon segment classification was presented in Table \ref{tab:evaluation}. From these results we can observe that the classification performance dropped significantly.

\section{Conclusion}
In this study, we proposed a novel camera localization system for colonoscopy videos. The localization system starts with the removal of non-informative frames and those containing biopsy forceps. The remaining frames are then fed into the motion estimation network to estimate camera motion between consecutive frames. The network is self-trained and does not require ground truth annotation. After that, the camera trajectory is derived, from which the location index is estimated. Based on the location index, a colon template was constructed by manually annotating times the camera entered each colon segment. With the estimated location index and colon template built from the training data, anatomical colon segment classification can be performed on a new colonoscopy video. The output of the localization system - the location index and anatomical colon segment classification - can facilitate further contextual understanding in automated colonoscopy video analysis. The anatomical disease severity distribution and biopsy site distribution can also be derived using this methodology, which can help evaluate a patient's condition and response to treatment. 

The algorithm was trained using colonoscopy videos from routine practice. The motion estimation network's performance was validated on an external dataset, with the results showing that the proposed method is more accurate than other published methods. The proposed localization system was also validated using colonoscopy videos from routine practice. The performance of the colon segment classification was calculated and compared with baselines using either the time index or ScopeGuide length. The results show that using a camera-based location index achieves the best performance in colon segment classification. Additionally, we compared the trajectories of the camera motion-based location index and the ScopeGuide length-based location index. Similar patterns can be observed when there is no sharp spike or bump in ScopeGuide length-based location index sequences. These results indicate that the proposed localization system can accurately determine location awareness.  

One limitation of this study is that the time at which the camera was withdrawn was manually annotated. This is due to the high proportion of non-informative frames during the insertion phase. The information loss resulting from frame removal reduces the accuracy of the camera localization during the insertion phase and makes it challenging to automatically identify the time at which the camera was withdrawn. In our future work, an image classifier will be trained to detect frames in the ileum or cecum based on textural features. Combined with the motion estimation results, the time at which the camera was withdrawn will be identified, which can lead to a fully-automated localization system. The proposed anatomical colon segment classification is limited by assuming the patient's colon and segments are of a normal length. In our future work, additional information, including surgical history, and automatically detected anatomical features (e.g., the appendiceal orifice) will be integrated to better identify the colon segment. The scale-drifting problem is an intrinsic limitation of the self-supervised camera pose estimation network. Our experimental results in Figure \ref{fig:tragectory} indicate that the scale-drifting problem is minor in the proposed system. In our future work, we will explore the possibility of integrating other information in the localization system. Though the colon surface is relatively uniform, some anatomical features may be detected in the cecum, transverse colon, and rectum. It is possible to build a method that leverages the information from colon segment templates and anatomical feature detection to refine the location index estimation for a new colonoscopy video.


\bibliographystyle{elsarticle-harv}\biboptions{authoryear}
\bibliography{reference}

\appendix
\setcounter{figure}{0}    
\setcounter{table}{0}

\section{Camera model and image distortion correction}
\label{sec:camera_model}
In this study, the camera at the tip of the colonoscope has a fisheye lens (Olympus PCF-H190). The fisheye lens achieves a very wide viewing angle by projecting a point in the 3D world frame to a half-hemisphere and then mapping the point to an image plane. 

\begin{figure}[] 
\includegraphics[width=3.5in]{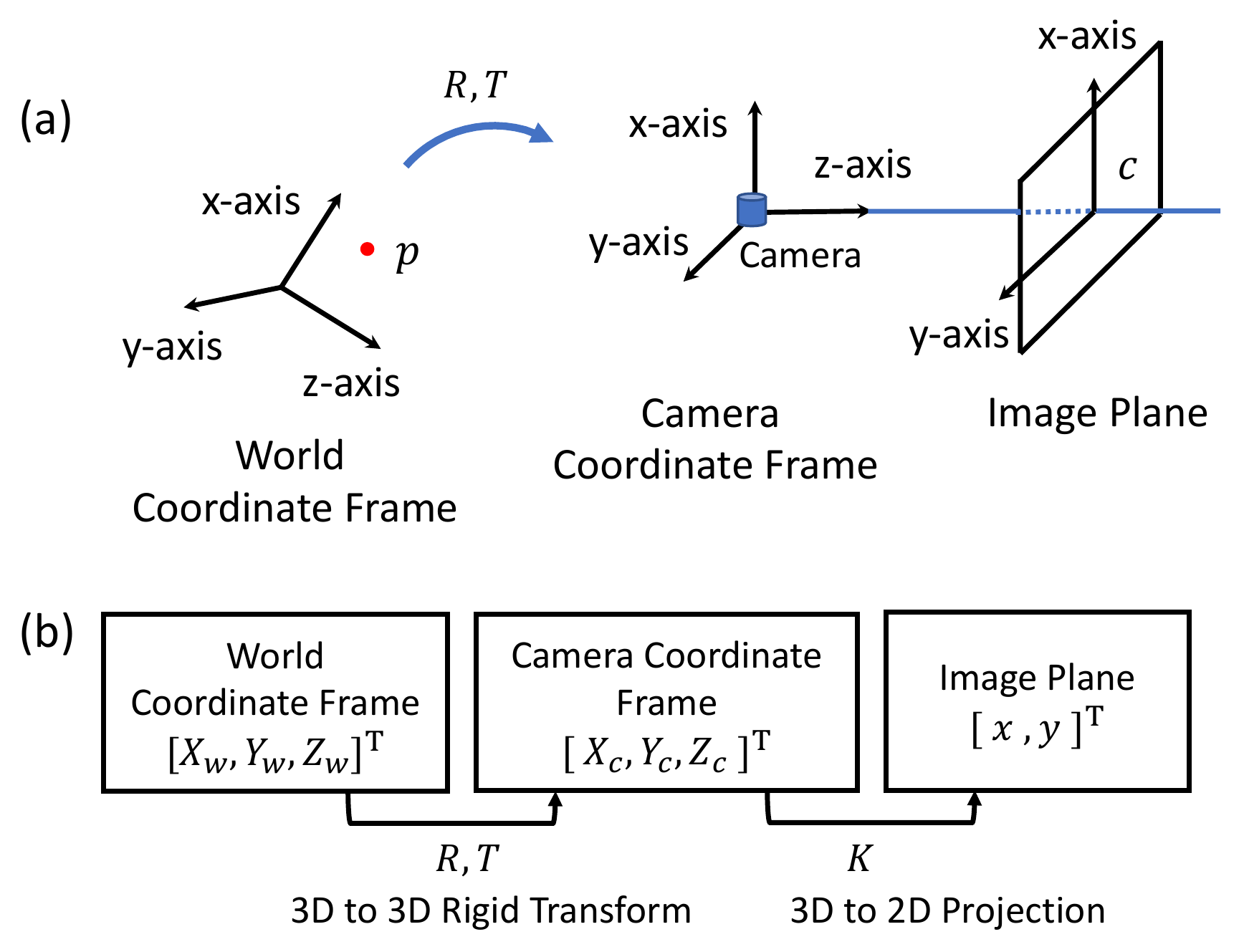}
\caption{Camera imaging model. (a) Diagram of each coordinate system. $p$ is a point in the 3D world. $c$ is the origin of the image plane (it is also called the image center); (b) A point in the world coordinate frame can be transformed into the camera coordinate frame via translation and rotation.  The point can then be projected onto the image plane of the camera.}
\label{fig:camera_model}
\end{figure}

Figure \ref{fig:camera_model} is a diagram of the camera imaging model. Figure \ref{fig:camera_model}(a) illustrates how a world coordinate frame, camera coordinate frame, and image plane are defined. The world coordinate frame is an arbitrarily-defined 3D coordinate system. The camera coordinate system is a 3D coordinate system based on the camera's optical center. The image plane is a 2D coordinate system, where the frame is generated from the camera, and its origin is the intersection between the $z$-axis (optical axis) and the image plane.

As shown in Figure \ref{fig:camera_model}(b), a point in the 3D world coordinate frame $[X_w, Y_w, Z_w]\tran$ can be transformed into the camera coordinate system using the extrinsic parameters
\begin{equation}
\begin{pmatrix} X_c
\\ Y_c
\\ Z_c
\end{pmatrix} = R \begin{pmatrix} X_w
\\ Y_w
\\ Z_w
\end{pmatrix} + T,
\label{equ:world-to-camera}
\end{equation}
where $R$ is a $3\times3$ rotation matrix and $T$ is a $3\times1$ translation vector. $R$ and $T$ are extrinsic parameters of a camera that represent the camera's location in the world coordinates and the camera's orientation with respect to the world coordinate axes. If the world frame is the same as the camera frame, $R$ is an identity matrix and $T$ is the zero vector.

The resulting point $[X_c, Y_c, Z_c]\tran$ in the camera coordinate frame can be projected onto the image plane as $[x, y]\tran$ by the camera. 

For a pinhole camera, the homogeneous coordinates $[x, y, 1]\tran$ of the projected point in the image plane can be written as
\begin{equation}
Z_c\begin{pmatrix} x
\\ y
\\ 1 \end{pmatrix}
= K\begin{pmatrix} X_c
\\ Y_c
\\ Z_c \end{pmatrix}, \hspace{0.1cm} K=\begin{pmatrix}
f_x & e & x_0 \\ 
0 & f_y & y_0\\ 
0&  0 & 1  
\end{pmatrix},
\label{equ:camera-to-image}
\end{equation}
where $K$ is called the intrinsic matrix of the camera, $f_x$ and $f_y$ are the focal lengths, $x_0$, $y_0$ are the principal point offsets, and $e$ is the axis skew.

In this study, the mathematical model of a fisheye camera proposed in \citep{scaramuzza2006toolbox} was utilized. The mapping of camera coordinates $[X_c, Y_c, Z_c]\tran$ to the 2D pixel coordinates $[x, y]\tran$ in the image plane of a fisheye camera can be written as

\begin{subequations}
\begin{align}
& \beta\begin{pmatrix} x-x_0
\\ y-y_0
\\ d(l)
\end{pmatrix} =
\beta\begin{pmatrix} x-x_0
\\ y-y_0
\\ \sum_{i=0}^n a_il^i
\end{pmatrix}
= \begin{pmatrix} X_c
\\ Y_c
\\ Z_c
\end{pmatrix} \\
& d(l) = \sum_{i=0}^n a_il^i \\
& l=\sqrt{(x-x_0)^2+(y-y_0)^2},
\end{align}
\label{equ:fisheye}
\end{subequations}
where $\beta$ is the scaling factor; $d(r)$ is the image distortion; $a_0$, $a_1$, ..., $a_n$ are intrinsic parameters of the camera; and $l$ is the distance from the image center $[x_0, y_0]\tran$.

With a fisheye lens the generated image will have a convex non-rectilinear appearance. Image distortion correction is essential for the subsequent optical flow calculation and camera motion estimation. For distortion correction, camera calibration is first performed to estimate the camera's intrinsic parameters $a_0$, ..., $a_n$, ($n$ can also be determined during camera calibration), and $K$. In the process of camera calibration, a paper with a checkerboard pattern is placed in front of the fisheye camera and a number of frames from different angles are captured. After that, the camera's intrinsic parameters are estimated by calculating how straight lines in the checkerboard are distorted. After the camera calibration, the distorted pixel coordinates $[x, y]\tran$ can be converted to pixel coordinates $[x', y']\tran$ using equations (\ref{equ:camera-to-image}) and (\ref{equ:fisheye}):
\begin{equation}
K^{-1}\begin{pmatrix} x'
\\ y'
\\ 1 \end{pmatrix} = \beta\begin{pmatrix}  x-x_0
\\ y-y_0
\\ d(r)
\end{pmatrix}.
\label{equ:conversion}
\end{equation}

Based on equation (\ref{equ:conversion}), $x$ and $y$ can be written as

\begin{subequations}
\begin{align}
x' &= \frac{f_x}{d(r)}(x-x_0)+\frac{e}{d(r)}(y-y_0)+x_0',\\ 
y' &= \frac{f_y}{d(r)}(y-y_0)+y_0',
\end{align}
\end{subequations}
where $[x_0', y_0']\tran$  is the center of the corrected image.
In this way, image distortion can be eliminated, and the corrected image can be regarded as an image captured by a pinhole camera with the estimated intrinsic matrix $K$.

\begin{figure}[!ht] 
\centering
\includegraphics[width=3.5in]{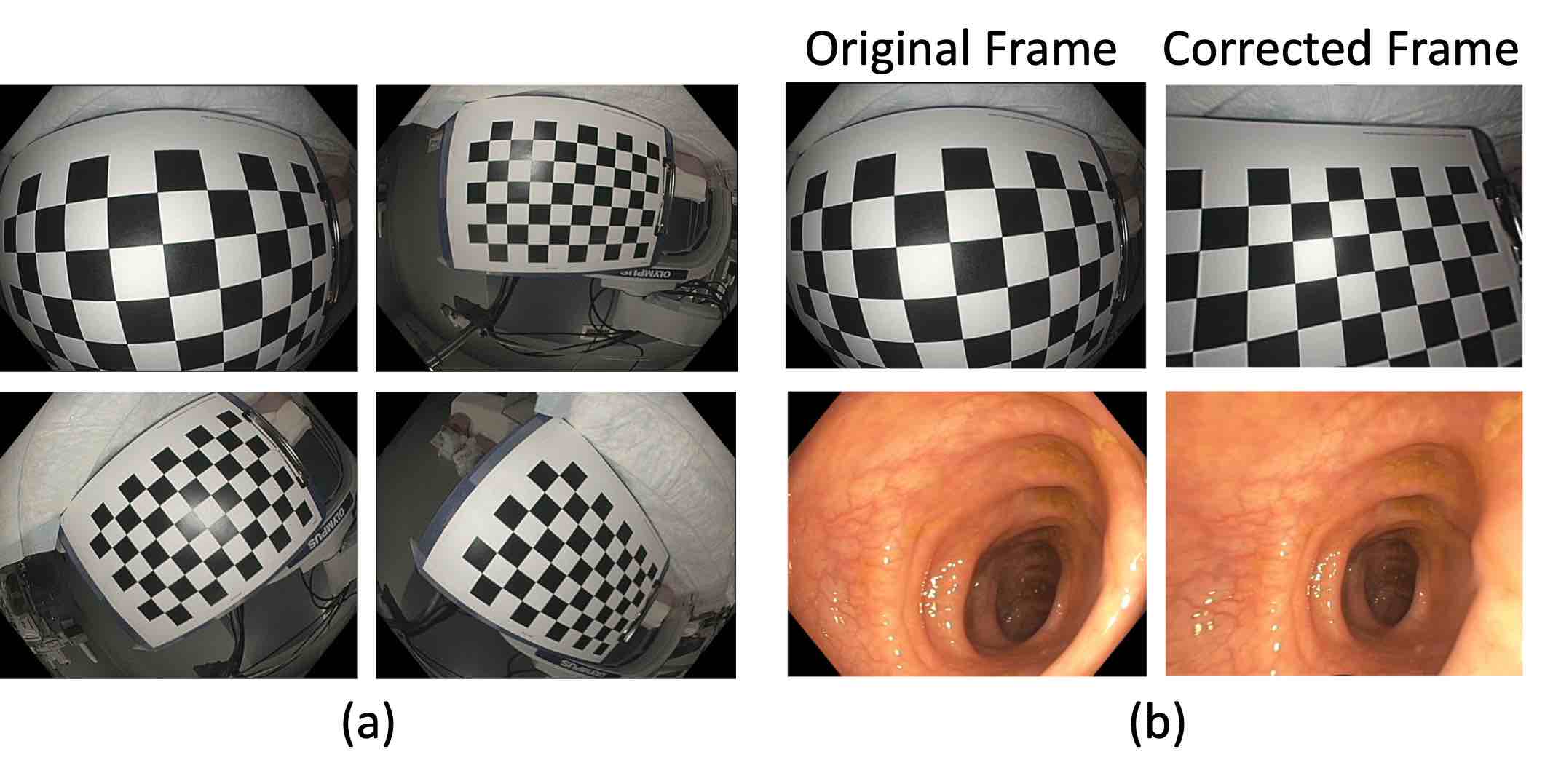}
\caption{Camera calibration and frame correction. (a) Examples of frames captured in camera calibration (b) Examples of image distortion correction.}
\label{fig:camera_calibration}
\end{figure}

Figure \ref{fig:camera_calibration} (a) presents several examples of frames captured for camera calibration. Frames of a checkerboard pattern were taken from different angles and distances to capture the characteristics of the fisheye lens. With the calculated intrinsic parameters, the distorted images can be corrected. Figure \ref{fig:camera_calibration} (b) shows two examples of distorted image correction. The first one is a frame of a checkerboard pattern,in which we can see the lines of the checkerboard pattern are straight after the original frame was corrected. The second one is a frame of the colon. All frames in colonoscopy videos from the localization dataset were corrected to remove distortion.

\section{Optical flow calculation}
\label{ap:optical_flow}
The proposed method requires optical flow as part of input to the motion estimation network. In this study, a pre-trained PWC-Net model from \citep{sun2018pwc} was used. The PWC-Net model was first trained on the FlyingChairs dataset consisting of 22,872 image pairs \citep{dosovitskiy2015flownet}  and then fine-tuned on the FlyingThings3D dataset consisting of 35,000 image pairs \citep{mayer2016large}. In their published results, PWC-Net achieved good performance with respect to several benchmarks. A deep learning based optical flow calculation method was chosen considering the complex geometry and limited textural pattern across the colon. From \citep{sun2018pwc}, PWC-Net is robust to real images where the image edges are often corrupted by motion blur and noise. We also evaluated the built-in Lucas-Kanade optical flow method from OpenCV library. On the validation set of the localization dataset Set 1, the performance of the pose estimation decreased slightly.

\section{Architectures of the motion network and disparity network}
\label{sec:a.architecture}

\begin{figure}[!ht]
\centering
\includegraphics[width=3.2in]{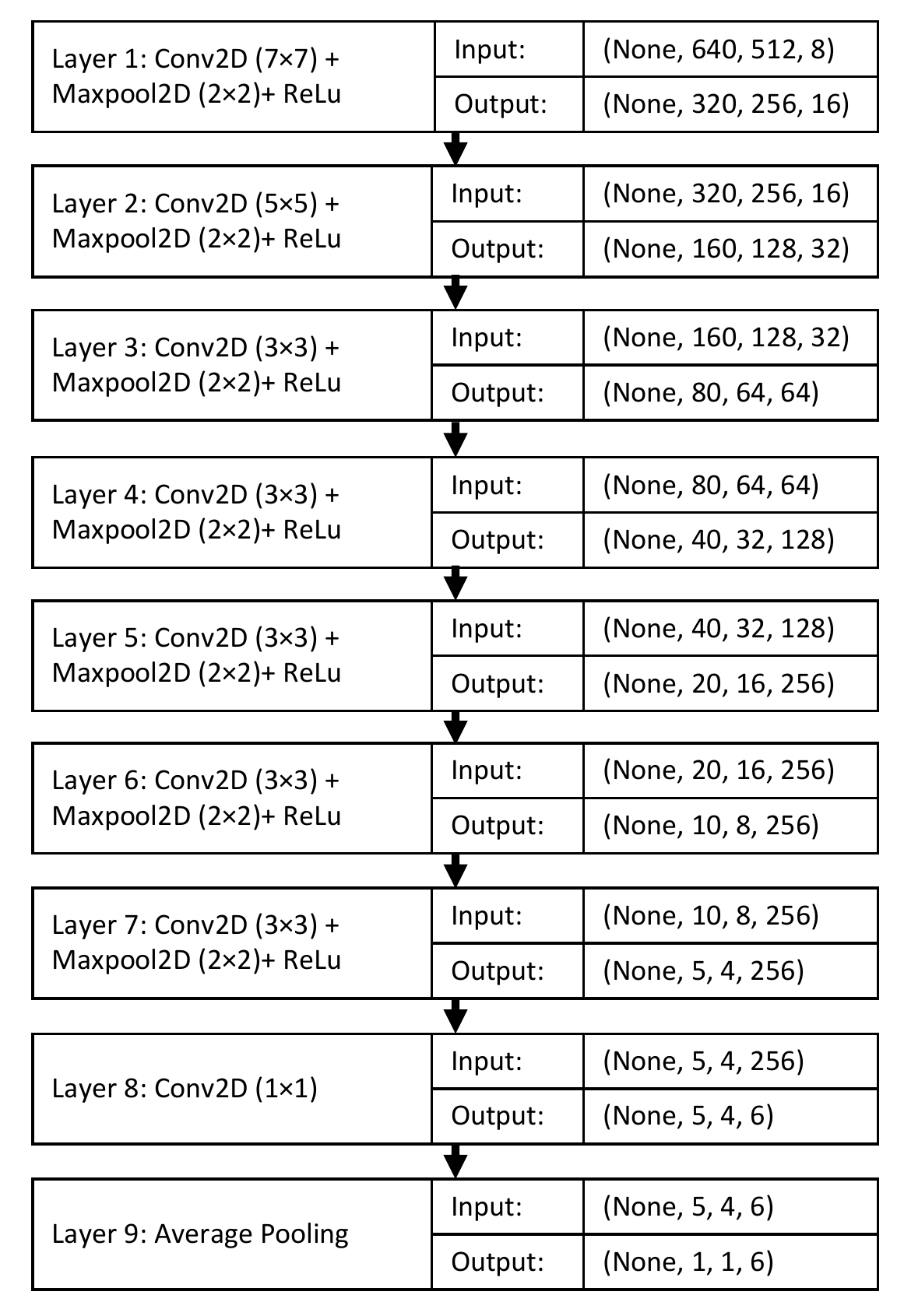}
\caption{The detailed architecture for the motion network.}
\label{fig:a.network_architecture_motion} 
\end{figure}

\begin{figure*}[hb]
\centering
\includegraphics[width=6.4in]{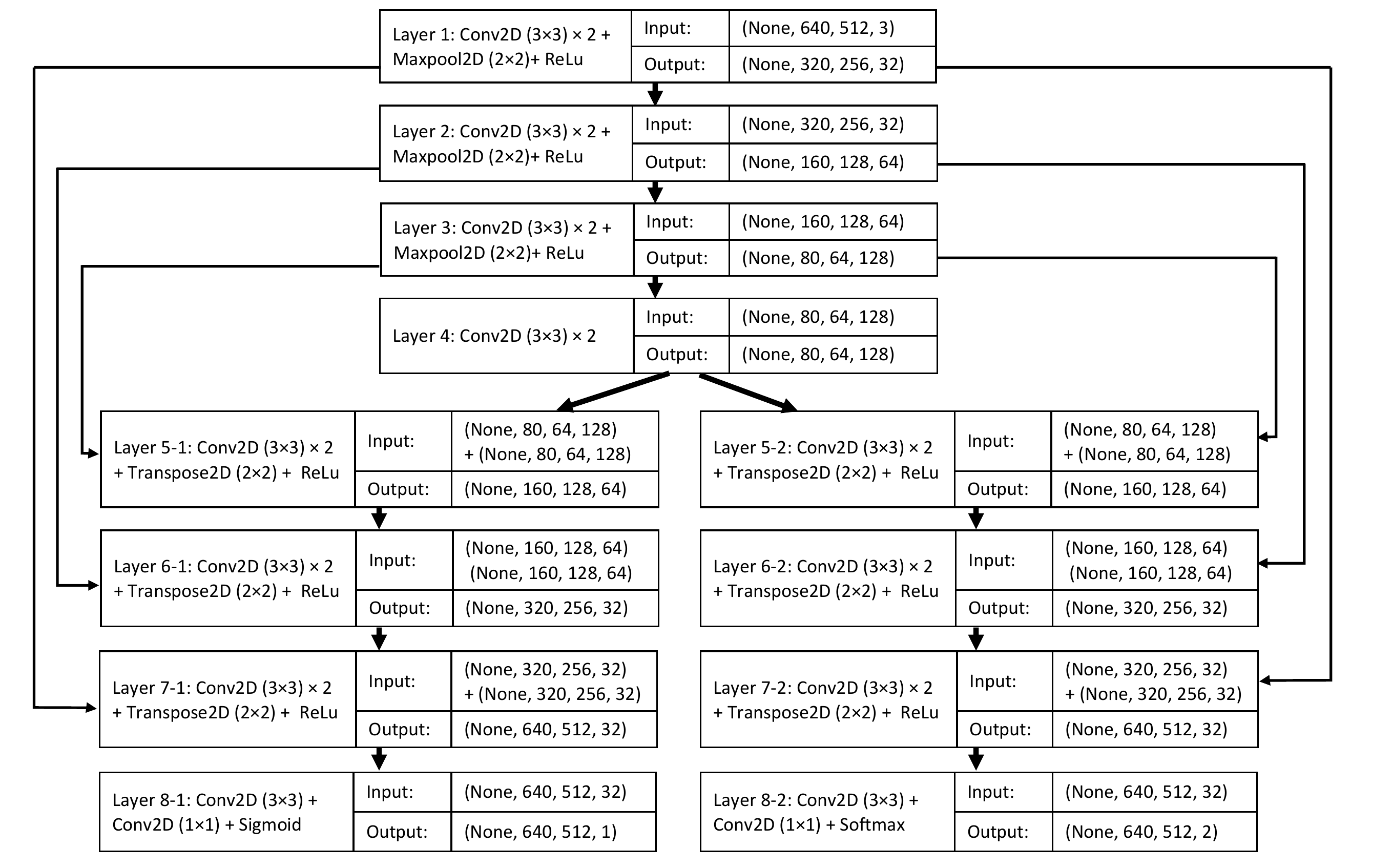}
\caption{The detailed architecture for the disparity network.}
\label{fig:a.network_architecture_disparity}
\end{figure*}

The detailed architectures of the motion network and disparity network are shown in Figure \ref{fig:a.network_architecture_motion} and Figure \ref{fig:a.network_architecture_disparity}. For each layer, the filter size, activation function, input size, and output size are given. In Figure \ref{fig:a.network_architecture_motion} and Figure \ref{fig:a.network_architecture_disparity}, ``Conv2D" means 2D convolutional layer; ``Maxpool2D" means 2D max pooling layer; ``Transpose2D" means 2D transpose layer. The shape of the input and output for each block is presented in the format of (batch size, height, width, the number of channels). ``None'' means that the batch size is of arbitrary value.

\section{Baseline using ScopeGuide length}
\label{appendix:scopeguide}
Currently, ScopeGuide is the best clinical measurement tool with FDA approval that can be used to estimate the traveled distance in colonoscopy videos. ScopeGuide's principle use is as a training tool to inform physicians of the shape and configuration of the scope inside the patient, not as a positioning system. The raw data from the electromagnetic sensors utilized by ScopeGuide is not accessible. The distance measurement provided, though crude, offered the best objective localization that is currently available and FDA approved for in human use.  Alternative methods would require real-time fluoroscopy (exposing the patient to ionizing radiation) or additional sensor-based approaches that are not FDA approved. Ideally, a specialized sensor would provide the best estimate of motion and spatial data, and its accuracy is posited to be superior to a vision-based system. However, vision-based camera localization methods are still attractive as there is no additional equipment needed, allowing for rapid integration into clinical workflows, wide availability, and low cost.

\section{Generic colon template and colon length variation}
In this study, we estimated a patient-generic template for the lengths of the colon segments to perform anatomical colon segment classification. There are some other sources that can be used to estimate the colon length. For example, in \citep{sadahiro1992analysis}, the colon segment length and variation across patients were analyzed for patients undergoing barium enema examination. CT or MR abdominal images can also be used to calculate a colon template. However, during a colonoscopy the colon will be stretched and the length of the colon segment will be different from the length under normal conditions. Additionally, the shapes of colon segments vary - the ascending and transverse colon are relatively straight, while the the sigmoid colon has an ‘S’ shape.   Considering the motility of the colon, its extreme distensibility, and the impact of patient position (different for colonoscopy, CT, or MRI), the shape, length, and configuration of the colon changes during colonoscopy.  As a result, the length of the colon segment after stretching cannot be inferred without a colonoscopy. Thus, CT or MR abdominal images cannot be used for colon segment template building. As the clinical importance of the localization algorithm is to link disease features on frames from colonoscopy video to location awareness, it is better to use the colon template built from colonoscopy videos. 

The colon template was constructed based on the assumption that the relative lengths of the colon segments are similar across patients. \citep{sadahiro1992analysis} shows the variation of absolute colon segment length across the patients.  Although the definition of colon segments is slightly different and the length cannot be used for building the colon template directly for the aforementioned reasons, the variation can be used as a reference. From their results, the variations in the length of the rectum and cecum are small while the variations in the length of the ascending colon to the sigmoid colon are relatively larger (up to 10 cm). In Table \ref{tab:colon_segment_variation}, the range and variation of the estimated length for each manually annotated colon segment on the independent test set (Set 3 of the localization set) are presented. Compared to the variation in \citep{sadahiro1992analysis}, the variation shown in Table \ref{tab:colon_segment_variation} is close and slightly higher, which may be due to the errors from location index estimation and manually colon segment annotations. Though variation of the relative colon segment length across patients exists, the anatomical colon segment classification is clinically important. Considering the anatomical geometry and the motility and distensibility of the colon, building a patient-generic template from colonoscopy videos is the best way of proceeding compared to other alternatives without introducing additional sensors. Many clinical applications only need relative location and generalize anatomic spatial information. For example, information like “approximately 20\% is affected and the region is near the sigmoid colon and rectum” can be very important to evaluate the patient’s condition and predict outcomes.

\begin{table*}[]
\centering
\small
\begin{tabular}{@{}ccccccc@{}}
\toprule
                   & Cecum & Ascending colon & Transverse colon & Descending colon & Sigmoid colon & Rectum \\ \midrule
Mean               & 0.067 & 0.142           & 0.245            & 0.204            & 0.244         & 0.097  \\
$25^{th}$ percentile    & 0.039 & 0.110            & 0.150             & 0.134            & 0.161         & 0.061  \\
$75^{th}$ percentile    & 0.087 & 0.171           & 0.336            & 0.284            & 0.306         & 0.103  \\
Standard deviation & 0.038 & 0.061           & 0.097            & 0.088            & 0.086         & 0.057  \\ \bottomrule
\end{tabular}
\caption{Average, range, and variation of the estimated length for each manually annotated colon segment on Set 3 of the localization set.}
\label{tab:colon_segment_variation}
\end{table*}

\end{document}